%% file: acl_latex.tex
\pdfoutput=1

\documentclass[11pt]{article}

\usepackage[]{acl}

\usepackage{times}
\usepackage{latexsym}
\usepackage{svg}
\usepackage{adjustbox}   

\usepackage{multirow}

\usepackage[T1]{fontenc}

\usepackage[utf8]{inputenc}

\usepackage{microtype}
\usepackage{subcaption}
\usepackage{inconsolata}
\usepackage{amsmath,amssymb,amsfonts}
\usepackage{multicol}
\usepackage{multirow}
\usepackage{booktabs}
\usepackage{graphicx}
\usepackage{array}
\usepackage{arydshln}
\usepackage{longtable}

\newcommand{\nwilds}{four}

\title{On the Robust Approximation of ASR Metrics}

\author{
  Abdul Waheed$^{1}$\,
  ~~~~~Hanin Atwany$^{2}$\,
  ~~~~~Rita Singh$^{1}$\,
    ~~~~~Bhiksha Raj$^{1}$\\
  $^{1}$Carnegie Mellon University~~~~~ 
  $^{2}$MBZUAI \\
  \texttt{\normalsize \{abdulw,bhiksha,rsingh\}@cs.cmu.edu}~~~~~~~~\texttt{\normalsize hanin.atwany@mbzuai.ac.ae}
}

\begin{document}
\maketitle
\begin{abstract}
Recent advances in speech foundation models are largely driven by scaling both model size and data, enabling them to perform a wide range of tasks, including speech recognition. Traditionally, ASR models are evaluated using metrics like Word Error Rate (WER) and Character Error Rate (CER), which depend on ground truth labels. As a result of limited labeled data from diverse domains and testing conditions, the true generalization capabilities of these models beyond standard benchmarks remain unclear. Moreover, labeling data is both costly and time-consuming. To address this, we propose a novel label-free approach for approximating ASR performance metrics, eliminating the need for ground truth labels. Our method utilizes multimodal embeddings in a unified space for speech and transcription representations, combined with a high-quality proxy model to compute proxy metrics. These features are used to train a regression model to predict key ASR metrics like Word Error Rate (WER) and Character Error Rate (CER). We experiment with over 40 models across 14 datasets representing both standard and in-the-wild testing conditions. Our results show that we approximate the metrics within a single-digit absolute difference across all experimental configurations, outperforming the most recent baseline by more than 50\%.

\end{abstract}

\input{sections/1_introduction}

\input{sections/2_related_work}

\input{sections/3_methodology}

\input{sections/4_experiments}

\input{sections/5_results}

\input{sections/6_conclusion}

\input{sections/7_limitations}

\input{sections/8_ethics}

\bibliography{custom}

\appendix

\input{sections/9_appendix}
\label{sec:appendix}


\end{document}

%% file: sections/1_introduction.tex
\section{Introduction}

Automatic Speech Recognition (ASR) models have made significant advancements in recent years, achieving near-human performance on several standard evaluation benchmarks~\citep[\textit{inter alia}]{radford2022robustspeechrecognitionlargescale, seamless2023, communication2023seamlessm4tmassivelymultilingual, Harper_NeMo_a_toolkit}. These models are typically evaluated using metrics like Word Error Rate (WER) and Character Error Rate (CER)~\cite{likhomanenko2020rethinking}, which are essential for assessing model performance.

However, these metrics are dependent on ground truths, which are often scarce in resource-constrained environments, and human labeling is both costly and time-consuming. To mitigate this challenge, several reference-free evaluation methods are proposed~\cite{yuksel2023noreferreferencelessqualitymetric, 7178922, Swarup2019, 9413966, 8325512, 5947648}. While these approaches eliminate the reliance on labeled data, they primarily offer relative assessments of transcription quality, rather than providing precise error counts or rates. As a result, their applicability in real-world scenarios, where actionable performance metrics are crucial for further model refinement and deployment, is limited.

Given the limitations of both methods, approximating ASR metrics has emerged as a promising alternative for label-free evaluation~\cite{chowdhury2023multilingualworderrorrate, sheshadri2021werbertautomaticwerestimation, ali-renals-2018-word}. This approach typically involves training regression~\cite{jalalvand-etal-2016-transcrater} and/or classification models~\cite{sheshadri-etal-2021-wer} on top of speech and text encoders. While this method offers a close approximation of error metrics, several important questions remain unresolved. Specifically, an approximation model trained on dataset sampled from \(D\) to predict ASR metrics for a source model \(M\) must be evaluated under diverse conditions: 1) on test data that is IID (independent and identically distributed) sampled from \(D\); 2) on out-of-distribution (OOD) data representing diverse domains and recording conditions; 3) on IID data, but transcription from a target model \(T\); and 4) on OOD data with transcriptions from a target model \(T\). Most prior works~\cite{chowdhury2023multilingualworderrorrate, sheshadri2021werbertautomaticwerestimation} focus primarily on the first condition. Moreover, recent advancements in multimodal foundation models offer new opportunities to directly train regression models on unified speech and text embeddings.

To address these critical research gaps, we propose a novel framework for approximating the performance of a wide range of ASR models, both on standard benchmarks and in-the-wild scenarios. Specifically, we compute the similarity between speech and text embeddings in a unified space, capturing the semantic alignment between the two modalities. Additionally, we incorporate a high-quality reference model as a proxy, based on the intuition that agreement with a reliable proxy correlates with transcription quality, as shown in prior works~\cite{waheed2025udistilwhisperlabelfreedatafiltering}. These features are then used to train a regression model to predict key ASR metrics, such as WER, CER, and absolute word and character error counts.

In summary, our work represents one of the most comprehensive studies to date on approximating ASR metrics at scale, in terms of both data and model coverage. Our proposed approach serves as a reference-free evaluation particularly suited for label-scarce scenarios. Beyond evaluation, our method is especially valuable for tasks such as pseudo-labeling, where high-quality transcriptions are essential for downstream applications like knowledge distillation~\cite{waheed-etal-2024-distill, gandhi2023distilwhisper}.

Our contributions are as follows:
\begin{itemize}
    \item We evaluate over 40 ASR models across 14 diverse evaluation setups, including both standard benchmarks and domain-specific, unseen conditions, followed by training regression models to approximate ASR metrics.
    \item We compare our approach with the most recent work on approximating ASR metrics and show over a 50\% reduction in absolute difference against the strong baseline.
    \item We conduct a rigorous ablation study to analyze the impact of different experimental configurations, providing deeper insights into the robustness of our approach. Our findings show that our method is resilient to diverse evaluation setups and requires only a small amount of training data.
\end{itemize}

\noindent\textbf{Outline.} The remainder of this paper is organized as follows: Section~\ref{sec:related_work} reviews related work. Section~\ref{sec:methodology} presents our proposed methodology. Sections~\ref{sec:experiments} and~\ref{sec:results} detail our experimental setup, results, and ablation study, respectively. Section~\ref{sec:conclusion} concludes the paper and outlines future directions.


%% file: sections/2_related_work.tex
\section{Related Work}\label{sec:related_work}
Automatic speech recognition (ASR) has witnessed significant advancements in recent years, primarily due to the scaling of both data and model size~\cite{radford2022robustspeechrecognitionlargescale, communication2023seamlessm4tmassivelymultilingual}. Transformer~\cite{vaswani2023attentionneed} based models, in particular, have significantly contributed to these developments by effectively capturing long-range dependencies and contextual nuances in speech, achieving state-of-the-art (SOTA) performance across diverse benchmarks~\cite{kheddar2024automatic,dhanjal2024comprehensive, zimerman2023long}. While traditional evaluation metrics like Word Error Rate (WER) and Character Error Rate (CER) are de-facto evaluation metrics in benchmarking ASR systems~\cite{lin2021speech, park2024character}, scenarios where ground truth transcriptions are unavailable have caught interest in reference-free ASR evaluation methods~\cite{karbasi2022asr, wang2024no, kuhn2024measuring}.

Reference-free ASR evaluation methods aim to estimate ASR performance without requiring ground truth transcriptions~\cite{ospanov2024towards}. Earlier approaches rely on heuristic features or metadata such as speaker demographics, background noise, and linguistic characteristics \cite{litman-etal-2000-predicting, yoon10b_interspeech}, limiting their applicability across varied contexts. However, recent advancements focus on deep learning-based frameworks, such as convolutional neural networks (CNNs)~\cite{elloumi2018asrperformancepredictionunseen} and contrastive learning methods~\cite{yuksel2023referencelessqualitymetricautomatic}, to predict ASR quality directly from encoded speech and text. For instance, methods like NoRefER~\cite{yuksel2023noreferreferencelessqualitymetric} use Siamese architectures fine-tuned on ASR hypotheses, achieving high correlation with traditional metrics and improving WER by optimizing hypothesis ensembling \cite{park2024character}.

Efforts to approximate ASR metrics explore hybrid approaches that combine traditional and reference-free methods, such as leveraging word confidence scores, linguistic embeddings, or post-processing adaptations to estimate WER and CER without explicit references~\cite{Ali2020WordER, ali-renals-2018-word, kuhn2024measuring, negri-etal-2014-quality}. However, these approaches often suffer from reliance on specific ASR models or domain characteristics, limiting their generalizability. Unlike existing methods, our work addresses these limitations by introducing a robust, model and data-agnostic framework that evaluates ASR outputs across diverse datasets and configurations, emphasizing adaptability to unseen domains and variations. 


%% file: sections/3_methodology.tex
\section{Methodology}\label{sec:methodology}

We present a scalable and robust method to approximate ASR performance metrics using multimodal unified embeddings, proxy references, and regression models. The primary goal is to eliminate reliance on ground-truth labels, enabling performance evaluation in label-scarce scenarios. The pipeline consists of three components: representation similarity in a unified speech-text embedding space, agreement with a \texttt{high-quality} proxy reference, and a regression model trained on these features to predict ASR metrics. Our pipeline diagram is shown in Figure~\ref{fig:pipeline_diagram}.

\subsection{Similarity in Unified Representation Space}
The foundation of our approach is the SONAR model~\cite{duquenne2023sonar}, a state-of-the-art multimodal (speech-text) model trained to produce unified embeddings for both speech and text inputs. Let $x_{\text{speech}}$ represent the input speech signal and $x_{\text{text}}$ denote the corresponding ASR-generated transcription. SONAR maps these inputs to a shared embedding space, generating $e_{\text{speech}}$ and $e_{\text{text}}$:
\begin{equation}
    e_{\text{speech}} = f_{\text{SONAR}}(x_{\text{speech}}),\quad e_{\text{text}} = f_{\text{SONAR}}(x_{\text{text}})
\end{equation}
where $f_{\text{SONAR}}$ represents the embedding model. The alignment between these embeddings is quantified using cosine similarity:
\begin{equation}
    \text{Similarity}(x_{\text{speech}}, x_{\text{text}}) = \frac{e_{\text{speech}} \cdot e_{\text{text}}}{\|e_{\text{speech}}\| \|e_{\text{text}}\|}
\end{equation}
The similarity metric serves as an indicator of transcription quality, with higher values suggest better alignment between speech and text representations.

\subsection{Agreement with a Proxy Reference}
To complement the similarity score, we utilize proxy references generated by a high-quality ASR model, denoted as $x_{\text{proxy}}$. The comparison between the ASR-generated transcription $x_{\text{text}}$ and the proxy reference $x_{\text{proxy}}$ is quantified using Word Error Rate (pWER) and Character Error Rate (pCER) as defined in Appendix~\ref{appsubsec:methodology}.


These metrics assess transcription quality by comparing it with a reliable proxy reference, without using ground-truth labels at any stage. Proxy references are dynamically selected by profiling 41 models across datasets and ranking them by average performance. For each target ASR model, the reference is the highest-ranking model other than the target itself. For example, if \texttt{whisper-large-v3} ranks highest, the reference for \texttt{whisper} will be the second-best model. This ensures the proxy reference is both relevant and reliable for evaluating the target model.

\input{figures/pipeline}

\subsection{Regression Model for Metric Prediction}
The extracted features, including similarity scores and proxy metrics, are concatenated to form the input to a regression model. Let $z = [\text{Similarity}, \text{pWER}/\text{pCER}]$ represent the feature vector. The regression model $g$ estimates the ASR metrics $\hat{y}$, denoted as aWER and/or aCER:
\begin{equation}
    \hat{y} = g(z)
\end{equation}

The regression model is an ensemble of Random Forest, Gradient Boosting, and Histogram-based Gradient Boosting regressors. Each base model is fine-tuned via grid search for hyperparameter optimization. The ensemble is trained to minimize the mean absolute error between predicted and ground-truth metrics. Additionally, a ridge regression model with non-negativity constraints is included in the ensemble to ensure predictions remain within valid ranges. Additional details of our regression pipeline are provided in Section~\ref{sec:experiments}, with hyperparameter details in Appendix~\ref{appsubsec:experiments}.

\subsection{Evaluation}\label{subsec:evaluation}
We evaluate the regression model's performance across four setups, including IID and OOD data and different model configurations. Specifically, we train our regression model on one ASR system (source) on one dataset and evaluate it on both IID and OOD data for the source and target models. We provide a detailed analysis of the distribution shift in Appendix~\ref{appsubsec:domain_shift}.

Let $\mathcal{D}_{M,B}$ denote the 10 benchmark datasets, and $\mathcal{D}_{M, W}$ represent the {\nwilds}~in-the-wild datasets, as described in Section~\ref{subsec:datasets}, where $M \in \{S, T\}$ refers to either the source model $S$ or the target model $T$. 

The regression model is trained on data $\mathcal{D}_{S,B}^{\text{train}} \sim \mathcal{D}_{S,B}$ and evaluated on the IID test set $\mathcal{D}_{S,B}^{\text{test-IID}} \sim \mathcal{D}_{S,B}$, consisting of $80\%$ and $20\%$ of the data, respectively. Additionally, the model is evaluated on $\mathcal{D}_{T,B}^{\text{test-IID}}$, $\mathcal{D}_{S, W}$, and $\mathcal{D}_{T, W}$. Below, we detail the formulation of each evaluation setup.

\paragraph{Case 1: IID Evaluation (Source \(S\))}\label{para:case_1_data_iid_eval}
The regression model is trained on $\mathcal{D}_{S,B}^{\text{train}}$ and evaluated on $\mathcal{D}_{S,B}^{\text{test-IID}}$. Let \(x_1^S = f(s, o^S)\) represent the similarity between input speech \(s\) and the ASR output \(o^S\), and \(x_2^S = g(o^S, r)\) represent the agreement with the proxy reference \(r\), where \(o^S\) is the ASR output produced by the source model \(S\). The evaluation is formulated as:
\begin{equation}
\mathcal{L}_{\text{IID}}^S = \mathbb{E}_{(x_1^S, x_2^S, y) \sim \mathcal{D}_{S,B}^{\text{test-IID}}} \big[\mathcal{L}(h(x_1^S, x_2^S), y)\big]
\end{equation}
 
\paragraph{Case 2: IID Evaluation (Target \(T\))}\label{para:case_2_data_iid_eval}
The regression model trained on $\mathcal{D}_{S,B}^{\text{train}}$ is evaluated on the IID test set $\mathcal{D}_{T,B}^{\text{test-IID}}$. Let \(x_1^T = f(s, o^T)\) represent the similarity between input speech \(s\) and the ASR output \(o^T\), and \(x_2^T = g(o^T, r)\) represent the agreement with the proxy reference \(r\), where \(o^T\) is the ASR output produced by the target model \(T\). The evaluation is expressed as:
\begin{equation}
\mathcal{L}_{\text{IID}}^T = \mathbb{E}_{(x_1^T, x_2^T, y) \sim \mathcal{D}_{T,B}^{\text{test-IID}}} \big[\mathcal{L}(h(x_1^T, x_2^T), y)\big]
\end{equation}

\paragraph{Case 3: OOD Evaluation (Source \(S\))}\label{para:case_3_data_ood_eval}
The regression model trained on $\mathcal{D}_{S,B}^{\text{train}}$ is evaluated on the out-of-distribution set $\mathcal{D}_{S, W}$. Let \(x_1^S = f(s, o^S)\) represent the similarity between the input speech \(s\) and the ASR output \(o^S\), and \(x_2^S = g(o^S, r)\) represent the agreement with the proxy reference \(r\), where \(o^S\) is the ASR output produced by the source model \(S\). The evaluation is defined as:
\begin{equation}
\mathcal{L}_{\text{OOD}}^S = \mathbb{E}_{(x_1^S, x_2^S, y) \sim \mathcal{D}_{S, W}} \big[\mathcal{L}(h(x_1^S, x_2^S), y)\big]
\end{equation}

\paragraph{Case 4: OOD Evaluation (Target \(T\))}\label{para:case_4_data_ood_eval}
The regression model trained on $\mathcal{D}_{S,B}^{\text{train}}$ is evaluated on the out-of-distribution set $\mathcal{D}_{T, W}$, using the ASR output produced by the target model \(T\). Let \(x_1^T = f(s, o^T)\) represent the similarity between the input speech \(s\) and the ASR output \(o^T\), and \(x_2^T = g(o^T, r)\) represent the agreement with the proxy reference \(r\), where \(o^T\) is the ASR output produced by the target model \(T\). The evaluation is expressed as:
\begin{equation}
\mathcal{L}_{\text{OOD}}^T = \mathbb{E}_{(x_1^T, x_2^T, y) \sim \mathcal{D}_{T, W}} \big[\mathcal{L}(h(x_1^T, x_2^T), y)\big]
\end{equation}

\noindent\textbf{Note.} For computational feasibility, the primary experiments train the regression model on 9 out of the 10 datasets in $\mathcal{D}_{S,B}^{\text{train}}$ and evaluate it on the remaining dataset, as well as on all {\nwilds} datasets in $\mathcal{D}_{S,B}^{\text{OOD}}$. This process is repeated for each dataset in $\mathcal{D}_{S,B}^{\text{train}}$, ensuring robust evaluation across various testing conditions. No examples from $\mathcal{D}_{M,\text{OOD}}$ are used at any stage for training the regression model.

%% file: figures/pipeline.tex
\begin{figure}[!ht]
    \centering
    \includegraphics[width=\linewidth]{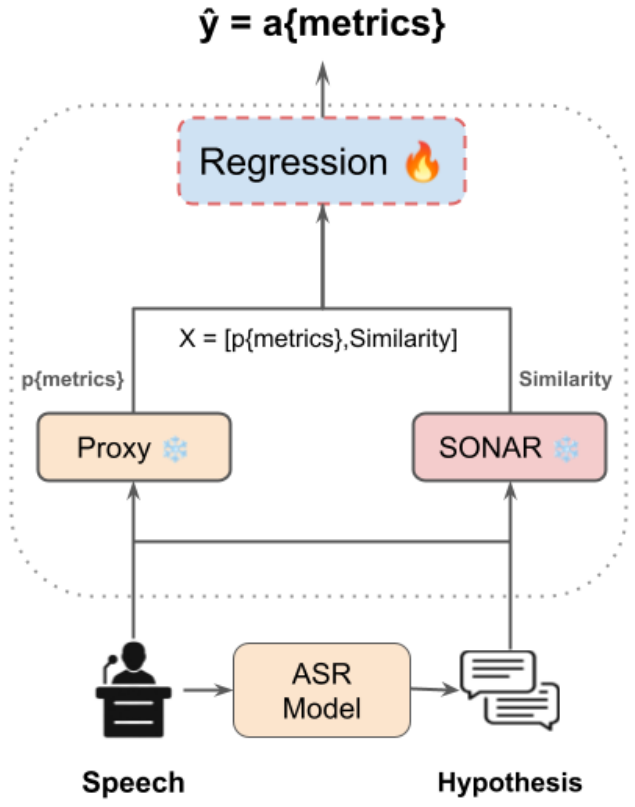}
    \caption{High-level diagram for our framework. The proxy is an ASR model that takes input speech and generates a transcription. We use the output from the source model as a hypothesis, and the output from the proxy model as a reference, to calculate metrics like WER and CER (p{metrics}), which we denote as pWER and pCER. We then use this, along with the the similarity between SONAR embeddings of the input speech and the hypothesis, to train the regression that gives approximated metrics (a{metrics}), e.g., aWER/aCER.}
    \label{fig:pipeline_diagram}
\end{figure}

%% file: sections/4_experiments.tex
\section{Experiments}\label{sec:experiments}
In this section, we present the experimental setup to evaluate our ASR metrics approximation tool. We describe the datasets, models, and regression pipeline used in our experiments, highlighting the diversity of ASR systems and testing conditions. 

\subsection{Datasets}\label{subsec:datasets}
To evaluate the robustness and generalizability of our ASR metrics approximation tool, we use datasets sourced from multiple distributions, divided into two types: \textbf{Standard Benchmark} and \textbf{Wild Challenge} datasets. We describe these datasets below and provide additional details in Appendix~\ref{appsubsec:datasets}, Table~\ref{tab:list_of_datasets}.

\noindent\textbf{Standard Benchmark Datasets.} We include widely used datasets representing diverse domains and acoustic conditions. \textit{LibriSpeech}~\cite{panayotov2015librispeech} provides 1,000 hours of English read audiobooks, covering both clean and noisy conditions. \textit{TED-LIUM}~\cite{rousseau2014tedlium} consists of TED talks from 2,000 speakers. \textit{GigaSpeech}~\cite{chen2021gigaspeech} spans audiobooks, podcasts, and YouTube, incorporating both read and spontaneous speech. \textit{SPGISpeech}~\cite{kensho2021spgispeech} features 5,000 hours of earnings calls with a focus on orthographic accuracy. \textit{Common Voice}~\cite{ardila2020common} is a multilingual, crowdsourced corpus with diverse accents. \textit{Earnings22}~\cite{rio2022earnings} provides 119 hours of accented, real-world earnings calls. Additional datasets include \textit{AMI (IHM)}~\cite{ami2005corpus}, with 100 hours of English meeting recordings from non-native speakers, and \textit{People's Speech}~\cite{galvez2021peoplesspeechlargescalediverse}, emphasizing inclusivity and linguistic diversity. \textit{SLUE-VoXCeleb}~\cite{shon2022slue} contains conversational voice snippets, capturing diverse speaking styles and emotions.

\noindent\textbf{Wild Datasets.} The wild set focuses on real-world variability and challenging scenarios. \textit{Primock57}~\cite{papadopoulos-korfiatis-etal-2022-primock57} includes telemedicine consultations with diverse accents, ages, and scenarios, recorded by clinicians and actors. \textit{VoxPopuli Accented}~\cite{wang-etal-2021-voxpopuli} contains multilingual speeches from European Parliament recordings, rich in named entities. \textit{ATCOsim}~\cite{jan_van_doorn_2023} features 10 hours of non-native English speech from air traffic control simulations with clean utterance-level transcriptions. Additionally, we include a noisy subset of \textit{LibriSpeech}~\cite{panayotov2015librispeech}, which reflects challenging real-world conditions.

In addition to the above datasets, we also run a small experiment to assess the cross-lingual transferability of the trained regression model. Specifically, we train the model on English data and evaluate it on both German and English data, and vice versa, using the English and German splits from LibriSpeech~\cite{panayotov2015librispeech} and Common Voice~\cite{ardila2020common}. In our ablation study, we compare the trained regression model with a proxy reference. To broaden this evaluation, we expand the in-the-wild dataset to include four additional datasets: a privately collected Medical ASR dataset with clinical conversations; standard data with eight synthetic perturbations (\textit{white noise}, \textit{time stretch}, \textit{pitch shift}, \textit{cross-lingual noise}, \textit{reverberation}, \textit{pub noise}, \textit{echo}, and \textit{distortion}); noisy home recordings BERSt~\cite{tuttösí2025berstingscreamsbenchmarkdistanced}; and CHiME-6(noisy subset)~\cite{watanabe2020chime}.

\subsection{Models}\label{subsec:models}

We evaluate our approximation framework for a range of state-of-the-art ASR models, put into three categories based on their architecture and functionality. Below we describe the datasets and provide additional details in Appendix~\ref{appsubsec:datasets} and in Table~\ref{tab:list_of_models}.

\noindent\textbf{Encoder-Decoder Models.}  
We include multiple encoder-decoder families of models capable of performing ASR tasks in a zero-shot setting. More specifically, we include \texttt{whisper}~\cite{radford2023robust} and \texttt{distil-whisper}~\cite{gandhi2023distilwhisper} models that perform really well across diverse testing settings. We also include \texttt{seamless }~\cite{communication2023seamlessm4tmassivelymultilingual, seamless2023, Barrault2025}, \texttt{SpeechT5}~\cite{ao2022speecht5unifiedmodalencoderdecoderpretraining} which are unified encoder-decoder frameworks for tasks such as ASR, speech synthesis, translation, and voice conversion. \texttt{MMS}~\cite{pratap2023scalingspeechtechnology1000} supports hundreds of languages and excels in resource-constrained scenarios. \textit{Moonshine (2)}~\cite{jeffries2024moonshinespeechrecognitionlive}, a lightweight and efficient model, is designed for edge deployments with strong performance.
Additionally, we include speech language models like \textit{SpeechLLM}~\cite{Rajaa_SpeechLLM_Multi-Modal_LLM}, which combine speech embeddings with language models to predict metadata such as speaker attributes, emotions, and accents, offering robust multimodal capabilities.

\noindent\textbf{NeMo-ASR Models.}  
We use multiple models from the \texttt{NeMo-ASR}~\cite{gulati2020conformerconvolutionaugmentedtransformerspeech, variani2020hybridautoregressivetransducerhat, noroozi2024statefulconformercachebasedinference, tang2023hybridtransducerattentionbased, Harper_NeMo_a_toolkit} toolkit by NVIDIA. We include models such as Canary and Parakeet, which use highly efficient speech encoders like Fast-Conformer~\cite{rekesh2023fastconformerlinearlyscalable}. In addition to that, we use models based on various encoders and decoders (CTC, RNN-T, TDT, Conformer-CTC~\cite{guo2021multispeakerasrcombiningnonautoregressive}. In our work, we evaluate 11 models from the NeMo-ASR toolkit.

\noindent\textbf{Encoder-Only.}  
We include self-supervised encoder-only models and their derivatives. Specifically, we use \texttt{Wav2Vec2}~\cite{schneider2019wav2vecunsupervisedpretrainingspeech, baevski2020wav2vec20frameworkselfsupervised}, \textit{HuBERT}~\cite{hsu2021hubertselfsupervisedspeechrepresentation}, and \textit{Data2Vec}~\cite{baevski2022data2vecgeneralframeworkselfsupervised}.

\subsection{Experimental Setup}

We evaluate all models listed in Section~\ref{subsec:models} on $1000$ examples sampled randomly from the $test$ split of each dataset, as described in Section~\ref{subsec:datasets}. Since all models are trained at a 16 kHz sampling rate, we (re)sample the speech accordingly. For ASR, we use greedy decoding and all other parameters are default unless otherwise specified. We apply basic text post-processing~\footnote{\url{https://bit.ly/enormwhisper}} before computing ASR metrics. We obtain all models from \texttt{Huggingface Hub}~\footnote{\url{https://huggingface.co/models}} and implement the ASR pipeline using the Transformers~\cite{wolf2020huggingfacestransformersstateoftheartnatural} library.

For multimodal embeddings, we use SONAR~\cite{duquenne2023sonar}, a 1024-dimensional sentence-level multilingual model. Specifically, we utilize \texttt{text\_sonar\_basic\_encoder} for text encoding and \texttt{speech\_sonar\_basic\_encoder} for speech encoding. 

The regression framework uses a stacking ensemble with base regressors and a final estimator. Hyperparameter tuning is performed with \texttt{RandomizedSearchCV} to minimize MAE. The model is trained on 9 benchmark datasets and evaluated on the remaining benchmark dataset and {\nwilds}~in-the-wild datasets. This process is repeated for all 10 benchmark datasets. Additional details of the regression pipeline are provided in Section~\ref{sec:methodology} and low-level details in Appendix~\ref{appsubsubsec:regression_pipeline}. 

We conduct ASR experiments on a single A100/H100 GPU, while the regression model training runs on CPUs. Although ASR time and memory consumption depend on the model size, embedding extraction for $1000$ audio-text pairs takes approximately one minute on a single consumer-grade GPU without parallelization or additional efficiency measures. Appendix~\ref{appsubsec:experiments} provides further experimental setup details.

\noindent\textbf{Baselines.} 
Recent studies directly aligned with our approach are limited. For instance, eWER~\cite{ali-renals-2018-word} and eWER2~\cite{Ali2020WordER} estimate error rates based on the input signal, which differs from our approach. In contrast, we incorporate the model's output transcript into the error rate approximation function. The most closely related recent works are WERBERT~\cite{sheshadri-etal-2021-wer} and eWER3~\cite{chowdhury2023multilingualworderrorrate}, which share a similar pipeline. Both use encoders for text, speech, and other data, followed by a regression model trained in an end-to-end setting. Since eWER3 is the more recent of the two, we use it as our baseline. In eWER3, the speech encoder is \emph{wav2vec2}~\cite{baevski2020wav2vec20frameworkselfsupervised}, and the text encoder is \emph{roberta-base}~\cite{liu2019robertarobustlyoptimizedbert}, with a regression model trained on top while both encoders remain frozen. Given the unavailability of public code or pretrained models for evaluation, we implement eWER3 with some modifications to ensure a fair comparison. Specifically, we extract features from both encoders and apply PCA for dimensionality reduction on each modality before training our regression pipeline. For both speech and text, we experiment with 32 and 64 PCA components (referred to as $nc$ in Table~\ref{tab:ablation_baseline_results}).

%% file: sections/5_results.tex
\section{Results}\label{sec:results}

We conduct experiments using two dataset categories: standard benchmarks and in-the-wild, as described in Section~\ref{subsec:datasets}. For each ASR model, a leave-one-out strategy is used, training the regression model on 9 benchmark datasets and testing it on the remaining benchmark dataset and all {\nwilds}~in-the-wild datasets to ensure comprehensive evaluation on out-of-domain data. Additionally, in-domain testing is included in ablation studies, as detailed in Section~\ref{subsec:ablation}. The regression model is trained to predict absolute error counts (word and character levels), which are normalized by the reference length to compute approximate error rates (\(aWER\) and \(aCER\)). We also train regression models to directly predict WER and CER. We provide results for fine-grained metrics in Table~\ref{tab:finegrained_metrics}.

\subsection{Evaluation on In-the-Wild Datasets}\label{subsec:wild_eval}

The wild datasets provide a realistic testbed for evaluating the regression model's ability to approximate error rates under real-world conditions. As shown in Table~\ref{tab:main_results_in_the_wild}, high-performing models, like \textit{canary-1b}, demonstrate strong agreement between predicted and actual error rates. For example, on VP\_Accented, \textit{canary-1b} achieves mean absolute difference of \(1.1\%\). On Primock57, the model shows robustness with a WER of \(16.2\%\) and an \(aWER\) of \(13.4\%\), highlighting its effective generalization across diverse and domain-specific contexts.

For models like \textit{data2vec-audio-large-960h} our approximation is pretty close to actual error rates with difference consistently under \(2\%\) on various datasets. For example, on LibriSpeech-test-noise, the model's actual WER is \(7.2\%\) while the approximated \(aWER\) is \(8.6\%\). Even on acoustically complex datasets like ATCOsim, where the WER is \(44.0\%\) and the \(aWER\) is \(51.1\%\), the model exhibits a reasonable alignment between approximated and actual error rates.

In contrast, models with high actual error rates, such as \textit{mms-1b-fl102}, show slightly larger deviations, particularly on datasets with challenging conditions. For instance, on ATCOsim, the WER is \(93.4\%\) and the \(aWER\) is \(99.0\%\), resulting in a significant deviation of \(5.6\%\), the highest observed across all in-the-wild datasets. Similarly, on Primock57, where the WER is \(70.2\%\) and the \(aWER\) is \(67.8\%\), the approximation also struggles to align due to the inherently high error rates. This highlights that extreme error cases often correspond to semantically nonsensical outputs, where the distinction between high and extremely high error rates becomes less relevant.

\input{tables/results_in_the_wild}

\subsection{Evaluation on Benchmark Datasets}\label{subsec:benchmark_eval}
We summarize results on 10 standard benchmark datasets in Appendix~\ref{appsubsec:results} Tables~\ref{tab:results_benchmark_part1} and~\ref{tab:results_benchmark_part2}. Each table reports actual WER/CER alongside the approximated WER/CER (denoted by aWER/aCER).

Overall, models such as \textit{parakeet-tdt-1.1b} and \textit{whisper-large-v3} show relatively small differences between WER and aWER, indicating reliable approximations. For instance,  the actual WER for \textit{whisper-large-v3} on \textbf{AMI\_IHM} is 19.0\% compared to aWER 17.1\%, 1.9\% gap. Conversely, some challenging datasets (e.g., \textbf{CV11} and \textbf{Earnings22}) reveal larger discrepancies for specific models, particularly those with higher overall error rates. For example, \textit{mms-1b-fl102} exhibits a wide WER/aWER gap on \textbf{Earnings22}, suggesting difficulty handling accented or domain-specific speech.

In general, high-performing ASR models demonstrate small WER–aWER gaps, indicating that it's easy to approximate when error rates are low. However, models with higher WERs or faced with more acoustically or linguistically challenging test sets tend to show wider divergences. Despite these variations, most results remain within a reasonable margin, highlighting the robustness of our approximation model on diverse out-of-distribution data.

\input{figures/results-four-models}

These results underscore the critical role of model quality in achieving reliable approximations. The approximation framework remains effective for high-performing models, while deviations tend to increase in cases of semantically divergent or poorly structured outputs, reflecting the inherent challenges in approximating errors for low-performing systems.

\input{sections/multiligual_results}

\subsection{Ablation}\label{subsec:ablation}
We conduct comprehensive ablation experiments to evaluate the robustness of the approximation model and the contributions of its individual components. Using the evaluation setup outlined in Section~\ref{subsec:evaluation}, we select \textit{data2vec-audio-base-960h} as the source model (\(S\)) and \textit{wav2vec2-base-960h} as the target model (\(T\)). The results are summarized in Table~\ref{tab:ablation_baseline_results}, where IID results correspond to Case-I~\ref{para:case_1_data_iid_eval}, and \(D\), \(M\), and \(D+M\) under OOD represent Case II~\ref{para:case_2_data_iid_eval}, Case-III~\ref{para:case_3_data_ood_eval}, and Case-IV~\ref{para:case_4_data_ood_eval}, respectively. The reference model's $r$ value represents the average WER across all datasets. We include reference models with varying $r$ values, such as \textit{whisper-large-v3}~($r=17.8$), \textit{whisper-medium.en}~($r=20.1$), \textit{whisper-tiny}~($r=33.4$), and \textit{mms-1b-fl102}~($r=51.0$).

The results in Table~\ref{tab:ablation_baseline_results} demonstrate the importance of proxy references in improving the regression model's performance. Training without proxy references (\textit{w/o PR}) significantly increases the mean absolute error (MAE) across all conditions. For instance, the IID MAE increases from \(1.03\) (Base) to \(3.13\), and the OOD \(D+M\) MAE rises from \(1.07\) (Base) to \(3.33\), highlighting the essential role of proxy references in approximation.

Increasing the number of high-quality proxy references (\textit{MPR}) further reduces errors. Under IID conditions, the MAE decreases from \(1.00\) with \(n=2\) to \(0.93\) with \(n=5\). Similarly, in OOD \(D+M\), the error drops from \(1.06\) (\textit{MPR}, \(n=2\)) to \(0.95\) (\textit{MPR}, \(n=5\)), demonstrating that multiple high-quality references enhance model robustness.

\input{tables/ablation_baseline}

The quality of references, quantified by the \(r\)-value, also plays a critical role. For example, in IID conditions, the MAE increases from \(1.31\) for \(r=17.8\) to \(2.03\) for \(r=51.0\). A similar trend is observed in OOD \(D+M\), where the MAE rises from \(1.40\) (\(r=17.8\)) to \(2.09\) (\(r=51.0\)). The absence of similarity (\textit{w/o S}) combined with low-quality proxies further degrades performance, underscoring the importance of both high-quality references and similarity measures. We provide character-level error count approximation in Appendix~\ref{appsubsec:results} Table~\ref{tab:ablation_baseline_results_cer}.

\paragraph{Scaling Training Data for Regression.}\label{effect_of_scaling_training_data} To evaluate the impact of training data size on the regression model, we scale the data from 1K to 10K examples in increments of 1K. As shown in Figure~\ref{fig:scaling}, the model's performance does not exhibit a clear trend with increasing training data size. Some datasets show slight improvements with more data; others show minimal improvement. This suggests that the regression model is largely agnostic to the size of the training data. In fact, it appears that a relatively small dataset of just 1,000 examples is sufficient to train a robust approximation model. This underscores the model's ability to generalize effectively with limited data, making it an efficient choice for scenarios with constrained datasets.
\input{figures/scaling}

\paragraph{Direct Comparison with Proxy Reference.}
We evaluate our regression model and standard reference-free baseline that computes the word error rate between the target and a proxy hypothesis. We report the mean absolute difference and compare the regression model (\textsc{Ours}) with direct calculation with proxy (\textsc{w Proxy}) in Table~\ref{tab:direct_comparison_with_proxy}.

Our results show that the regression model consistently reduces the absolute difference compared to direct evaluation with the proxy across seven of the eight datasets, with the only exception being \textsc{MedicalASR}. We also find that the regression model trained using only similarity features remains competitive. These findings demonstrate that our approach generalizes well across domains and provides more reliable reference-free ASR quality estimates than simply computing the error against a proxy reference.

\input{tables/direct_comparision_with_proxy}



%% file: tables/results_in_the_wild.tex
\begin{table}[!hbt]
\centering
\Large
\resizebox{\columnwidth}{!}{%
\begin{tabular}{p{3cm}p{2cm}p{2cm}p{2cm}p{2cm}}
\toprule
Model              & LS\_Noise & Primock57 & ATCOsim     & VP\_Acc \\ \midrule
w2v2-ls            & 8.8/10.2  & 32.8/35.6 & 43.0/49.5   & 20.4/26.4    \\
can-1b             & 4.1/6.4   & 16.2/13.4 & 30.4/35.5   & 23.2/12.1    \\
d2v-base           & 14.9/16.4 & 39.6/41.7 & 66.0/71.2   & 28.4/33.8    \\
d2v-large          & 7.2/8.6   & 28.3/30.7 & 44.0/51.1   & 21.4/26.5    \\
distil-l-v2        & 7.3/9.2   & 18.3/13.0 & 69.5/66.7   & 14.9/14.5    \\
distil-l-v3        & 6.1/8.3   & 18.4/12.9 & 69.0/63.6   & 14.8/14.0    \\
distil-s.en        & 9.1/10.6  & 19.3/14.7 & 74.9/69.1   & 14.6/14.7    \\
sm4t-l             & 11.2/12.3 & 41.7/37.8 & 75.0/82.5   & 29.3/19.9    \\
sm4t-m             & 14.9/15.6 & 44.1/39.7 & 54.6/60.4   & 30.5/22.5    \\
hub-l-ls-ft        & 7.3/8.8   & 29.5/32.0 & 50.4/56.9   & 21.4/26.6    \\
hub-xl-ls-ft       & 6.8/8.3   & 31.1/32.9 & 46.7/53.0   & 21.8/27.7    \\
mms-1b-a           & 9.5/11.1  & 36.2/34.4 & 63.4/71.8   & 29.9/23.8    \\
mms-1b-f102        & 24.0/24.9 & 70.2/67.8 & 93.4/99.0   & 39.4/38.2    \\
moon-b             & 11.3/12.4 & 19.9/18.5 & 65.5/66.2   & 17.1/20.8    \\
moon-t             & 15.5/17.4 & 29.2/29.5 & 62.9/68.5   & 22.1/26.2    \\
par-ctc-0.6b       & 4.6/7.4   & 16.3/13.8 & 32.9/42.9   & 16.3/13.8    \\
par-ctc-1.1b       & 4.5/6.9   & 16.6/14.1 & 30.9/39.9   & 16.4/12.4    \\
par-rnnt-0.6b      & 3.8/6.9   & 16.3/13.2 & 31.6/41.8   & 17.3/12.6    \\
par-rnnt-1.1b      & 3.5/6.1   & 14.6/13.3 & 27.3/37.6   & 18.1/10.4    \\
par-tdt-1.1b       & 3.4/6.0   & 13.5/13.2 & 28.3/35.7   & 17.9/10.2    \\
pkt-ctc-110m       & 6.1/8.6   & 16.7/13.0 & 39.9/42.4   & 19.2/12.5    \\
sm4t-v2-l          & 7.2/8.4   & 34.6/31.7 & 52.4/57.6   & 33.8/24.5    \\
spchllm-1.5B       & 15.3/16.6 & 42.0/41.8 & 121.1/125.4 & 57.0/59.3    \\
spchllm-2B         & 13.9/15.6 & 39.4/40.3 & 60.6/64.1   & 39.2/44.1    \\
stt-cfc-l          & 5.8/6.8   & 16.1/17.6 & 35.9/38.0   & 18.6/11.5    \\
stt-cfc-s          & 9.7/11.2  & 22.2/24.6 & 43.7/47.7   & 16.4/15.6    \\
stt-fc-cfc-l       & 6.8/10.0  & 17.6/23.9 & 34.9/47.6   & 18.9/13.3    \\
stt-fc-td-l        & 6.0/8.8   & 17.0/20.6 & 34.5/46.5   & 21.1/15.1    \\
w2v2-960h          & 17.4/18.5 & 44.7/47.1 & 68.4/74.0   & 29.9/36.5    \\
w2v2-crelpos       & 5.9/7.4   & 28.5/30.3 & 47.2/54.0   & 22.4/26.7    \\
w2v2-crope         & 6.6/8.1   & 31.7/33.4 & 49.8/56.9   & 21.9/26.3    \\
w2v2-l-960h        & 11.6/12.6 & 37.8/40.2 & 66.4/72.7   & 26.3/33.3    \\
w2v2-l-lv60-s   & 7.8/9.4   & 33.1/35.5 & 40.5/48.8   & 19.3/24.9    \\
w2v2-l-rft-ls      & 10.0/11.5 & 32.2/34.6 & 48.9/55.7   & 22.0/28.6    \\
whisper-l          & 6.2/8.1   & 18.8/13.9 & 65.3/66.9   & 18.7/15.9    \\
whisper-l-v2       & 5.4/6.6   & 19.0/13.1 & 64.8/74.8   & 20.0/18.1    \\
whisper-l-v3       & 4.6/5.9   & 18.7/12.0 & 64.7/73.9   & 19.2/18.1    \\
whisper-l-v3-t & 4.9/6.0   & 18.5/12.3 & 66.0/72.5   & 24.3/23.2    \\
whisper-m.en       & 6.5/7.9   & 19.5/14.0 & 66.2/73.8   & 27.6/26.4    \\
whisper-s.en       & 8.2/9.7   & 20.0/15.1 & 67.1/73.8   & 17.3/17.5    \\
whisper-tiny       & 18.5/20.7 & 30.0/26.6 & 97.6/102.5  & 29.8/33.2   \\ 
\bottomrule
\end{tabular}
}
\caption{Actual and approximated WER ($\downarrow$), separated by a slash, on out-of-distribution wild datasets. The regression model is trained independently for each ASR model on standard benchmarks, making the wild datasets out-of-distribution. See Table~\ref{tab:main_results_in_the_wild_cer} for full names.}
\label{tab:main_results_in_the_wild}
\end{table}

%% file: figures/results-four-models.tex
    




\begin{figure}[h!]
    \centering
    \includegraphics[width=1.0\linewidth]{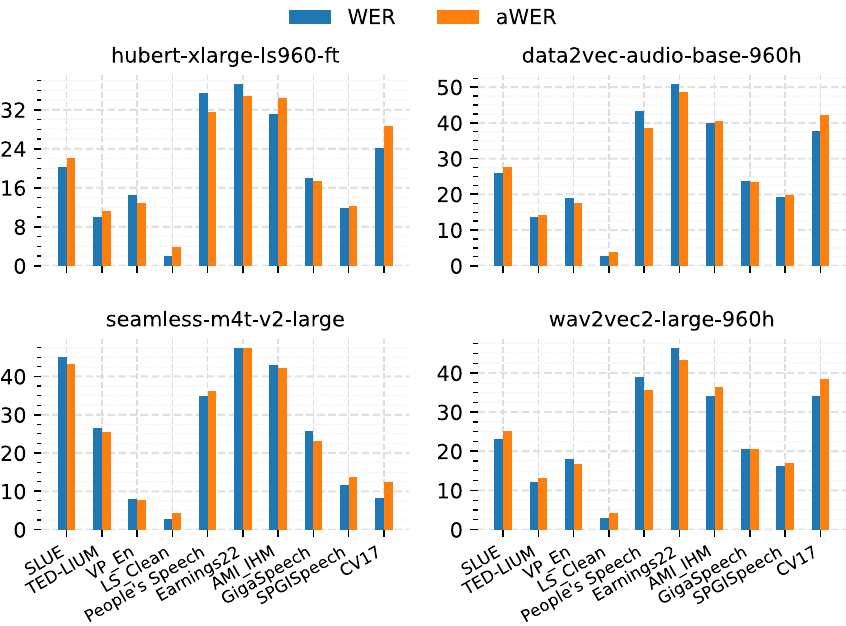}
    \caption{Actual and approximated WER for four models across standard benchmark.}
    \label{fig:fig-four-models}
\end{figure}

%% file: sections/multiligual_results.tex
\subsection{Multilingual Evaluation}\label{multilingual_evaluation}
We train the regression model on English and evaluate it on English and German, and vice versa. We do this experiment with two models as source and proxy, namely \textit{seamless-m4t-v2-large} and \textit{whisper-large-v3}. We report the mean absolute difference between approximated and actual word error counts in Table~\ref{tab:multilingual}.

\input{tables/multilingual_results}
We find that our framework demonstrates strong cross-lingual generalization: when trained on English data, the regression model maintains low absolute differences when evaluated on German datasets, and vice versa. This consistency across languages and datasets, using both \textit{seamless-m4t-v2-large} and \textit{whisper-large-v3} as source-proxy pairs, underscores the robustness and language-agnostic nature of our approach. These results validate that our method can be effectively applied in multilingual settings without the need for language-specific adaptations.

%% file: tables/multilingual_results.tex
\begin{table}[h]
\centering
\small
\renewcommand{\arraystretch}{1}
\setlength{\tabcolsep}{6pt}

\adjustbox{max width=\columnwidth}{%
\begin{tabular}{lcccc}
\toprule
\multicolumn{5}{c}{\textbf{Source:} \texttt{seamless-m4t-v2-large} \quad
                  \textbf{Proxy:} \texttt{whisper-large-v3}} \\
\midrule
Train$\setminus$Test & LS\_De & CV17\_De & LS\_En & CV17\_En \\
\midrule
LS\_De    & --   & 2.16 & 1.59 & 1.98 \\
CV17\_De  & 1.93 & --   & 0.50 & 0.56 \\
LS\_En    & 1.60 & 0.75 & --   & 0.66 \\
CV17\_En  & 1.82 & 0.66 & 0.56 & --   \\
\bottomrule
\end{tabular}}

\vspace{0.8em}

\adjustbox{max width=\columnwidth}{%
\begin{tabular}{lcccc}
\toprule
\multicolumn{5}{c}{\textbf{Source:} \texttt{whisper-large-v3} \quad
                  \textbf{Proxy:} \texttt{seamless-m4t-v2-large}} \\
\midrule
Train$\setminus$Test & LS\_De & CV17\_De & LS\_En & CV17\_En \\
\midrule
LS\_De    & --   & 2.03 & 1.11 & 1.96 \\
CV17\_De  & 1.74 & --   & 0.51 & 0.76 \\
LS\_En    & 1.29 & 1.50 & --   & 1.42 \\
CV17\_En  & 2.29 & 0.68 & 1.69 & --   \\
\bottomrule
\end{tabular}}

\caption{Cross-lingual mean absolute difference ($\downarrow$) between predicted and actual word error counts. Lower values mean better approximation.}
\label{tab:multilingual}
\end{table}

%% file: tables/ablation_baseline.tex

\begin{table}[!hbt]
\centering
\small 
\renewcommand{\arraystretch}{0.9} 
\setlength{\tabcolsep}{5pt} 
\begin{tabular}{lcccc}
\specialrule{1.2pt}{0.5em}{0.2em} 
\multirow{2}{*}{Method} & \multirow{2}{*}{IID} & \multicolumn{3}{c}{OOD}                  \\

\cmidrule(lr){3-5}

                        &                      & D        & M       & D + M \\ 
                        
\specialrule{1.2pt}{0.5em}{0.2em} 

eWER3(nc=32) & 2.03\textsuperscript{0.07} & 2.09\textsuperscript{0.04} & 2.06 \textsuperscript{0.03} & 2.12\textsuperscript{0.04} \\

eWER3(nc=64) & 1.98\textsuperscript{0.06} & 2.07\textsuperscript{0.05} & 2.00\textsuperscript{0.04} & 2.09\textsuperscript{0.05} \\

\midrule

Base                                    & 1.03\textsuperscript{0.03}          & 1.05\textsuperscript{0.01} & 1.03\textsuperscript{0.02} & 1.07\textsuperscript{0.01}  \\ 

\noalign{\vskip 0.3ex} 
\hdashline
\noalign{\vskip 0.6ex} 

w/o S                                 & 1.04\textsuperscript{0.03}          & 1.05\textsuperscript{0.01} & 1.04\textsuperscript{0.03} & 1.05\textsuperscript{0.01}  \\ 
w/o PR                                  & 3.13\textsuperscript{0.07}          & 3.22\textsuperscript{0.02} & 3.23\textsuperscript{0.05} & 3.33\textsuperscript{0.02}  \\ 

\noalign{\vskip 0.3ex} 
\hdashline
\noalign{\vskip 0.6ex} 

w/ MPR (n=2)                            & 1.00\textsuperscript{0.02}          & 1.04\textsuperscript{0.02} & 0.99\textsuperscript{0.02} & 1.06\textsuperscript{0.02}  \\
w/ MPR (n=3)                            & 0.96\textsuperscript{0.02}          & 0.97\textsuperscript{0.01} & 0.95\textsuperscript{0.02} & 0.99\textsuperscript{0.01}  \\
w/ MPR (n=4)                            & 0.95\textsuperscript{0.02}          & 0.96\textsuperscript{0.02} & 0.94\textsuperscript{0.02} & 0.98\textsuperscript{0.02}  \\
w/ MPR (n=5)                            & 0.93\textsuperscript{0.02}          & 0.93\textsuperscript{0.01} & 0.92\textsuperscript{0.02} & 0.95\textsuperscript{0.01}  \\ 

w/MPR (n=10) & 0.90\textsuperscript{0.02} & 0.93\textsuperscript{0.01} & 0.88\textsuperscript{0.02} & 0.95\textsuperscript{0.01} \\
w/MPR (n=20) & 0.89\textsuperscript{0.02} & 0.96\textsuperscript{0.02} & 0.87\textsuperscript{0.02} & 0.96\textsuperscript{0.02} \\

\noalign{\vskip 0.3ex} 
\hdashline
\noalign{\vskip 0.6ex} 

w/ mMPR (n=3)                           & 0.98\textsuperscript{0.02}          & 0.96\textsuperscript{0.02} & 0.97\textsuperscript{0.02} & 0.98\textsuperscript{0.02}  \\
w/ mMPR (n=5)                           & 0.94\textsuperscript{0.02}          & 0.94\textsuperscript{0.02} & 0.93\textsuperscript{0.01} & 0.96\textsuperscript{0.02}  \\ 

w/mMPR (n=10) & 0.92\textsuperscript{0.02} & 0.94\textsuperscript{0.02} & 0.91\textsuperscript{0.02} & 0.96\textsuperscript{0.02} \\
w/mMPR (n=20) & 1.04\textsuperscript{0.02} & 1.05\textsuperscript{0.01} & 1.02\textsuperscript{0.02} & 1.04\textsuperscript{0.01}
 \\

\noalign{\vskip 0.3ex} 
\hdashline
\noalign{\vskip 0.6ex} 

\multicolumn{1}{l}{Base (r=17.8)}    & 1.31\textsuperscript{0.04}          & 1.44\textsuperscript{0.02} & 1.31\textsuperscript{0.04} & 1.40\textsuperscript{0.01}  \\
\multicolumn{1}{l}{Base (r=20.1)}    & 1.36\textsuperscript{0.04}          & 1.36\textsuperscript{0.01} & 1.34\textsuperscript{0.03} & 1.34\textsuperscript{0.01}  \\
\multicolumn{1}{l}{Base (r=33.4)}    & 1.55\textsuperscript{0.04}          & 1.69\textsuperscript{0.02} & 1.55\textsuperscript{0.04} & 1.63\textsuperscript{0.02}  \\ 
\multicolumn{1}{l}{Base (r=51.0)}    & 2.03\textsuperscript{0.02}          & 2.10\textsuperscript{0.01} & 2.08\textsuperscript{0.05} & 2.09\textsuperscript{0.01}  \\ 

\noalign{\vskip 0.3ex} 
\hdashline
\noalign{\vskip 0.6ex} 

\multicolumn{1}{l}{w/o S (r=17.8)} & 1.47\textsuperscript{0.04}          & 1.56\textsuperscript{0.01} & 1.48\textsuperscript{0.04} & 1.54\textsuperscript{0.01}  \\
\multicolumn{1}{l}{w/o S (r=20.1)} & 1.55\textsuperscript{0.02}          & 1.50\textsuperscript{0.01} & 1.55\textsuperscript{0.03} & 1.50\textsuperscript{0.01}  \\
\multicolumn{1}{l}{w/o S (r=33.4)} & 1.79\textsuperscript{0.07}          & 1.89\textsuperscript{0.02} & 1.78\textsuperscript{0.06} & 1.82\textsuperscript{0.02}  \\
\multicolumn{1}{l}{w/o S (r=51.0)} & 2.23\textsuperscript{0.02}          & 2.24\textsuperscript{0.01} & 2.28\textsuperscript{0.04} & 2.21\textsuperscript{0.01} \\ 
\specialrule{1.2pt}{0.5em}{0.2em} 

\end{tabular}
\caption{Mean absolute error ($\downarrow$) between predicted word error count and actual error count (in absolute terms) across different configurations. PR - Proxy Reference, S - Similarity, MPR - Multiple PR, D - Data, M - Model. The OOD results are averaged across {\nwilds}~wild datasets. n is the number of proxy references. The r ($\downarrow$) value represents the average WER for proxy reference across 14 datasets.  Superscript represents the standard deviation across five runs.}
\label{tab:ablation_baseline_results}
\end{table}

%% file: figures/scaling.tex
\begin{figure}
    \centering
    \includegraphics[width=1.0\linewidth]{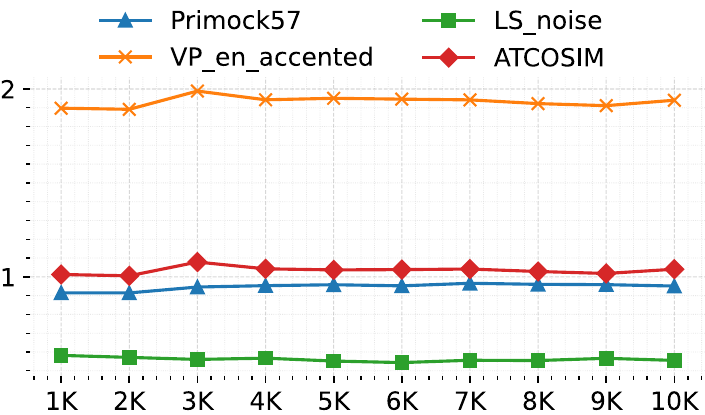}
    \caption{Mean absolute error (\(\downarrow\)) between predicted and actual word error counts across varying training data sizes for the regression model. The model was trained on 10 standard benchmarks and evaluated on {\nwilds}~in-the-wild test sets.}
    \label{fig:scaling}
\end{figure}



%% file: tables/direct_comparision_with_proxy.tex
\begin{table}[h]
\centering
\Large
\resizebox{\linewidth}{!}{%
\begin{tabular}{lccccc}
\toprule
\textbf{Dataset} & \multicolumn{2}{c}{PR (r=33.4)} & \multicolumn{2}{c}{PR (r=51.0)} & \textsc{w/o Proxy} \\
\cmidrule(lr){2-3}\cmidrule(lr){4-5}
 & \textsc{w Proxy} & \textsc{Ours} & \textsc{w Proxy} & \textsc{Ours} & \textsc{Ours} \\
\midrule
Primock57          & 2.25 & 1.54 & 4.89 & 2.63 & 3.82 \\
ATCOSim            & 5.35 & 2.11 & 3.56 & 2.02 & 2.64 \\
VP\_Accented       & 2.99 & 2.09 & 5.39 & 2.83 & 4.00 \\
LS\_Noise & 2.10 & 1.58 & 3.08 & 1.42 & 2.35 \\
BERSt              & 2.37 & 1.25 & 5.09 & 2.47 & 3.20 \\
Perturbed       & 1.84 & 1.34 & 3.38 & 1.75 & 2.81 \\
Medical         & 0.76 & 0.95 & 2.35 & 1.33 & 2.56 \\
CHiME6-Noisy       & 2.32 & 1.43 & 3.29 & 1.93 & 2.74 \\
\midrule
Average            & 2.50 & \textbf{1.54} & 3.88 & \underline{2.05} & 3.02 \\
\bottomrule
\end{tabular}}
\caption{Mean absolute difference ($\downarrow$) between the predicted and actual word error counts for our regression models and the baseline direct comparison with proxy reference (\textsc{w Proxy}).}

\label{tab:direct_comparison_with_proxy}
\end{table}

%% file: sections/6_conclusion.tex
\section{Conclusion}\label{sec:conclusion}
We present a framework for approximating ASR metrics, demonstrating its effectiveness in generalizing to unseen, in-the-wild, and challenging conditions. Our results show that the model performs well with absolute error counts, consistently outperforming strong baseline, with error rates remaining relatively low. We show that our proposed method achieves consistent performance across 40 ASR models and 14 evaluation setups, including both standard benchmarks and domain-specific conditions. The trained regression model can be efficiently used to approximate ASR metrics, particularly in data-constrained environments, such as critical domains with limited labeled data. In summary, our work bridges the gap between theoretical advancements and real-world applications, paving the way for more reliable and scalable ASR systems. While in this work, we 
evaluate monolingual and cross-lingual generalization,  
future work will focus on extending this framework to support a multilingual setting and exploring language-agnostic ASR metric approximation.

%% file: sections/7_limitations.tex
\section{Limitations}\label{sec:limitations}
In this work, we introduced a framework for approximating ASR metrics, evaluated across various ASR models and datasets. Despite the promising results, there are several limitations to consider.

\noindent\textbf{Evaluation.} While our evaluation setup is comprehensive, consisting of over 40 models and 14 datasets representing various acoustic and linguistic conditions such as natural noise, dialects, and accents—far surpassing previous works—we have not explored more nuanced conditions such as gender, non-native speech, and approximation across various age groups. Additionally, while the framework has shown strong performance in approximating ASR metrics across multiple datasets, its generalization to highly diverse or extreme real-world conditions might still require further investigation.

\noindent\textbf{Language.} Additionally, the current evaluation focuses solely on monolingual and bilingual settings. Extending this framework to include multiple languages and rigorously testing it across diverse linguistic contexts represents a critical direction for future research.

\noindent\textbf{Compute.} Unlike previous works, our final approximator is a simple regression model that does not require GPUs to run, we do utilize a single GPU for multimodal embedding extraction, which could be performed on any consumer-grade GPU.

%% file: sections/8_ethics.tex
\section{Ethics Statement}\label{sec:ethics_statement}

\paragraph{Data Collection and Release.} The datasets used in our experiments consist of publicly available ASR data from both benchmark and in-the-wild sources, as detailed in Section~\ref{subsec:datasets}. We ensure that the use of these datasets aligns with the principles of fair use, specifically in a non-commercial academic context or as specified in their original license. All datasets are openly accessible, and no private or confidential data is included in this work to the best of our knowledge. 

\paragraph{Intended Use.} By enabling the approximation of ASR performance metrics with minimal data, our work has the potential to impact applications in domains with limited data availability, such as healthcare, emergency response, and low-resource language research. We believe our approach will foster further research in scalable, low-cost ASR systems with comprehensive evaluation, benefiting industries and research areas that serve underrepresented or resource-limited populations.

\paragraph{Potential Misuse and Bias.} While our regression model has demonstrated effectiveness in approximating ASR metrics, it is important to consider potential misuse and bias. Given that our model is trained on diverse datasets, including those with various linguistic, acoustic, and demographic variations, there is a risk that the model may inherit biases present in the data, particularly with respect to accents, dialects, and socio-linguistic factors. Additionally, as our model approximates error rates, it could be misused in applications where the approximation may not be sufficient for real-world critical tasks. We recommend cautious deployment and further evaluation in sensitive applications, especially those where fairness and accuracy are critical.

%% file: sections/9_appendix.tex
\clearpage
\section{Appendix}\label{appsec:appendix}
\subsection{Methodology}\label{appsubsec:methodology}

\begin{align}
    \text{pWER}(x_{\text{text}}, x_{\text{proxy}}) &= \frac{\text{EditDistance}(x_{\text{text}}, x_{\text{proxy}})}{\text{WordCount}(x_{\text{proxy}})} \\
    \text{pCER}(x_{\text{text}}, x_{\text{proxy}}) &= \frac{\text{EditDistance}(x_{\text{text}}, x_{\text{proxy}})}{\text{CharCount}(x_{\text{proxy}})}
\end{align}

\subsection{Datasets}\label{appsubsec:datasets}

To evaluate the robustness and generalizability of our ASR metrics approximation tool, data were sourced from multiple repositories, which we divided into two distinct groups: Standard Benchmark and Wild Challenge dataset. 
\subsubsection{Standard Benchmark Datasets} 

There are six datasets in total that fall under the benchmark group. These datasets are categorized based on their frequent use in ASR model training and their representation of commonly encountered domains in real-world applications. 

\noindent\textbf{LibriSpeech~\cite{panayotov2015librispeech}.} prioritized speaker and content balance over explicit consideration of speech characteristics. It comprises approximately 1000 hours of English read audiobooks, with subsets featuring both clean and noisy speech conditions to simulate different acoustic environments. While the dataset covers diverse subject matter, its focus on formal, clear speech from public domain books means it lacks the natural variability of spontaneous speech, limiting its representation of conversational or informal dialogue.  \\
\noindent\textbf{TED-LIUM~\cite{rousseau2014tedlium}.} contains TED Talks totaling 452 hours of English speech data from approximately 2,000 speakers, recorded in close-talk microphone conditions. The corpus features narrated speaking styles, capturing clear and articulate speech. While it provides non-orthographic transcriptions, lacking formatting such as capitalization and punctuation, it remains a valuable resource for training and benchmarking automatic speech recognition (ASR) models.\\
\noindent\textbf{GigaSpeech~\cite{chen2021gigaspeech}.} is a multi-domain, multi-style speech recognition corpus incorporating diverse acoustic and linguistic conditions. It sources audio from three primary domains: audiobooks, podcasts, and YouTube, covering a wide range of speaking styles, including both read and spontaneous speech. The dataset covers a broad spectrum of topics, such as arts, science, sports, and more, making it highly versatile for training robust speech recognition models. \\
\noindent\textbf{SPGISpeech~\cite{kensho2021spgispeech}.} contains 5,000 hours of professionally transcribed audio from corporate earnings calls, featuring both spontaneous and narrated speaking styles. It emphasizes orthographic accuracy, providing fully formatted text with capitalization, punctuation, and denormalization of non-standard words. \\
\noindent\textbf{Common Voice~\cite{ardila2020common}.} (a multilingual corpus of narrated prompts built through crowdsourcing. Recorded in teleconference conditions, the corpus features narrated speaking styles and emphasizes inclusivity by covering a wide range of accents and languages, including low-resource ones. \\
\noindent\textbf{Earnings22~\cite{rio2022earnings}.} is a 119-hour corpus of English-language earnings calls from global companies, designed to address the lack of real-world, accented speech data in ASR benchmarking\\

\noindent\textbf{AMI (IHM)~\cite{ami2005corpus}.} The AMI Meeting Corpus is a 100-hour dataset of English meeting recordings, featuring multimodal data synchronized across close-talking and far-field microphones, room-view and individual cameras, slide projectors, and whiteboards. It includes mostly non-native speakers recorded in three rooms with varying acoustics. Digital pens capture unsynchronized handwritten notes, supporting research in speech recognition, diarization, and multimodal interaction. Available under edinburghcstr/ami, it is widely used for meeting analysis and speech processing studies. \\

\noindent\textbf{People's Speech~\cite{galvez2021peoplesspeechlargescalediverse}.}  Thousands of hours of labeled speech data collected from diverse speakers, covering a wide range of topics, accents, and speaking styles. The dataset emphasizes inclusivity and linguistic diversity, making it suitable for developing robust and generalized speech models. It is widely used in academic and industrial research to advance the state-of-the-art in automatic speech recognition (ASR) and other speech-related applications. \\

\noindent\textbf{SLUE - VolxCeleb~\cite{shon2022slue}.}consists of single-sided conversational voice snippets extracted from YouTube videos, originally designed for speaker recognition. The dataset represents natural, unscripted speech in diverse real-world settings, capturing a wide range of speaking styles, emotions, and acoustic conditions. Utterances containing slurs were excluded, and partial words were trimmed using a forced aligner to ensure clean, usable segments. \\

\subsubsection{Wild Challenge Set}

\noindent\textbf{Primock57~\cite{papadopoulos-korfiatis-etal-2022-primock57}.} 
contains mock consultations conducted by seven clinicians and 57 actors posing as patients, representing a diverse range of ethnicities, accents, and ages. Each actor was provided with a detailed case card outlining a primary care scenario, such as urinary tract infections, cardiovascular issues, or mental health concerns, ensuring the conversations were realistic and clinically relevant. The consultations were recorded using telemedicine software, capturing separate audio channels for clinicians and patients, and transcribed by experienced professionals to ensure accuracy.\\
\noindent\textbf{VoxPopuli Accented~\cite{wang-etal-2021-voxpopuli}.} is a comprehensive multilingual speech corpus derived from European Parliament event recordings. It includes audio, transcripts, and timestamps sourced directly from the official Parliament website. Due to its origin, the dataset features a rich collection of named entities, making it particularly suitable for tasks like Named Entity Recognition (NER). \\

\noindent\textbf{ATCOsim~}\cite{jan_van_doorn_2023}.is a specialized database containing ten hours of English speech from ten non-native speakers, recorded during real-time ATC simulations using close-talk headset microphones. It features orthographic transcriptions, speaker metadata, and session details. With a 32 kHz sampling frequency and 10,078 clean, utterance-level recordings. \\

\input{tables/datasets}
\subsection{Models}
\noindent\textbf{Whisper Models \cite{radford2023robust}.}
is a transformer-based model that processes 80-dimensional log-mel filter bank features from 16 kHz audio, utilizing a 2D CNN stack followed by a transformer encoder-decoder architecture. Trained on a vast multilingual dataset of 680,000 hours, it incorporates timestamp tokens into its vocabulary and operates on 30-second audio windows during inference, auto-regressively generating text sequences while leveraging encoder outputs as context. Variants of Whisper, such as Distilled, Large, Base, and Medium, offer different trade-offs in model size and performance, catering to diverse computational and accuracy requirements.

\noindent\textbf{Seamless Models~\cite{communication2023seamlessm4tmassivelymultilingual, seamless2023, Barrault2025}.}
is a cutting-edge multilingual and multitask model for speech and text translation. Built on the UnitY architecture, it uses w2v-BERT 2.0 for speech encoding and NLLB for text encoding, supporting nearly 100 languages. A text decoder handles ASR and translation, while a text-to-unit (T2U) model and multilingual HiFi-GAN vocoder generate speech. Leveraging SONAR embeddings and SeamlessAlign (443,000 hours of aligned speech/text data), it achieves SOTA results in ASR, speech-to-text, speech-to-speech, and text-to-text translation, excelling in low-resource languages. It introduces BLASER 2.0 for robust evaluation and outperforms competitors in noisy environments. \\

\noindent\textbf{Nemo-ASR-Models~\cite{gulati2020conformerconvolutionaugmentedtransformerspeech, variani2020hybridautoregressivetransducerhat, rekesh2023fastconformerlinearlyscalable, noroozi2024statefulconformercachebasedinference, tang2023hybridtransducerattentionbased,  Harper_NeMo_a_toolkit}} We included several NVIDIA’s NeMo advanced automatic speech recognition (ASR) models, including Canary, Parakeet (110M, 0.6B, and 1.1b), Conformer-CTC, and Fast-Conformer, as each is designed for specific use cases and optimized for performance. Canary-1B is a state-of-the-art multilingual, multitask model featuring a FastConformer encoder and Transformer decoder. The Parakeet family includes models with a FastConformer encoder paired with different decoders: CTC, RNN-T, or TDT. Conformer-CTC is a non-autoregressive model based on the Conformer architecture, combining self-attention and convolution for global and local feature extraction. It uses CTC loss and a linear decoder, supporting both sub-word (BPE) and character-level encodings. While Fast-Conformer is an optimized version of the Conformer architecture, offering significant speed improvements (2.4x faster) with minimal quality degradation. It uses 8x depthwise convolutional subsampling and reduced kernel sizes for efficiency. 

\noindent\textbf{Wav2Vec2 Models~\cite{schneider2019wav2vecunsupervisedpretrainingspeech, baevski2020wav2vec20frameworkselfsupervised}.} is a self-supervised pre-trained model designed to process raw audio inputs and generate speech representations. The model architecture consists of three key components: a convolutional feature encoder, a context network, and a quantization module.  The convolutional feature encoder converts raw waveforms into latent representations, which are then processed by the context network a transformer based stack with 24 blocks, a hidden size of 1024, 16 attention heads, and a feed-forward dimension of 4096 to capture contextual information.The quantization module maps these latent representations to quantized forms. 

\noindent\textbf{HuBERT Models~\cite{hsu2021hubertselfsupervisedspeechrepresentation}.}
 is a self-supervised learning framework designed for speech representation learning where CNN-encoded audio features are randomly masked. During training, the model predicts cluster assignments for masked regions of the input speech, forcing it to learn both acoustic and language models from continuous inputs.
 
\noindent\textbf{Audio/Speech Language Models 1.5B and 2B ~\cite{Rajaa_SpeechLLM_Multi-Modal_LLM}} is a multi-modal Language Model designed to analyze and predict metadata from a speaker's turn in a conversation. It integrates a speech encoder to convert speech signals into meaningful embeddings, which are then processed alongside text instructions by TinyLlama-1.1B-Chat-v1.0 to generate predictions. The model accepts 16 KHz audio inputs and predicts metadata such as SpeechActivity, Transcript, Gender, Age, Accent, and Emotion. 

\noindent\textbf{SpeechT5~\cite{ao2022speecht5unifiedmodalencoderdecoderpretraining}.}
unified modal framework capable of handling a wide range of tasks, including automatic speech recognition (ASR), speech synthesis, speech translation, voice conversion, speech enhancement, and speaker identification.Its audio post-net, which can incorporate speaker embeddings to enable prosody transfer, making it effective for tasks like voice conversion and speech synthesis. By leveraging its encoder-decoder architecture, SpeechT5 can generate high-quality mel-spectrograms from text input while preserving speaker-specific characteristics like emotion and gender.
\input{tables/models}

\subsection{Experiments}\label{appsubsec:experiments}

\subsubsection{Regression Pipeline.}\label{appsubsubsec:regression_pipeline}
The regression framework is a stacking ensemble comprising multiple base regressors and a final estimator. We perform basic hyperparameter tuning using \texttt{RandomizedSearchCV} with 5-fold cross-validation, with the objective to minimize \textit{mean absolute error (MAE)}. The search explores key hyperparameters such as \texttt{n\_estimators}, \texttt{max\_depth}, \texttt{learning\_rate}, and \texttt{min\_samples\_split}, balancing model complexity and generalization. We provide hyperparameter and other details in~\ref{tab:hyperparameters}. The model is trained on 14 datasets divided into two groups: \textit{bench} (10 standard benchmark datasets) and \textit{in-the-wild} (4 diverse, real-world datasets). A leave-one-out strategy is applied to the \textit{bench} set, where the model is trained on 9 datasets and evaluated on the remaining one. All trained models are also evaluated on the \textit{in-the-wild} set, which remains isolated during training to assess out-of-domain generalization.
\input{tables/hyperparameters}

\input{sections/distribution_shift}

\subsection{Results}\label{appsubsec:results}
\input{tables/ablation_results_appendix}
\input{tables/results_in_the_wild_cer}

\input{sections/finegrained_metrics}

\input{figures/all_models}

\input{tables/results_benchmark}

%% file: tables/datasets.tex
\begin{table*}[ht]
\renewcommand{\arraystretch}{1}
\centering
\resizebox{\linewidth}{!}{%
\begin{tabular}{p{4cm}p{2cm}p{8cm}}
\toprule
\textbf{Name} & \textbf{Type} & \textbf{Description} \\ \midrule
\textbf{LibriSpeech} & Bench & A corpus of approximately 1,000 hours of 16kHz read English speech, derived from LibriVox audiobooks, segmented and aligned for ASR tasks. \\
\textbf{TED-LIUM} & Bench & Contains TED Talks totaling 452 hours of English speech data from approximately 2,000 speakers, recorded in close-talk microphone conditions. \\
\textbf{GigaSpeech} & Bench & A multi-domain, multi-style speech recognition corpus incorporating diverse acoustic and linguistic conditions, sourced from audiobooks, podcasts, and YouTube. \\
\textbf{SPGISpeech} & Bench & Contains 5,000 hours of professionally transcribed audio from corporate earnings calls, featuring both spontaneous and narrated speaking styles. \\
\textbf{Common Voice} & Bench & A multilingual corpus of narrated prompts built through crowdsourcing, recorded in teleconference conditions, covering a wide range of accents and languages. \\
\textbf{Earnings22} & Bench & A 119-hour corpus of English-language earnings calls from global companies, designed to address the lack of real-world, accented speech data in ASR benchmarking. \\
\textbf{AMI (IHM)} & Bench & The AMI Meeting Corpus is a 100-hour dataset of English meeting recordings, featuring multimodal data synchronized across various devices. \\
\textbf{People's Speech} & Bench & Contains thousands of hours of labeled speech data collected from diverse speakers, covering a wide range of topics, accents, and speaking styles. \\
\textbf{SLUE - VoxCeleb} & Wild & Consists of single-sided conversational voice snippets extracted from YouTube videos, originally designed for speaker recognition. \\
\textbf{Primock57} & Wild & Contains mock consultations conducted by seven clinicians and 57 actors posing as patients, representing a diverse range of ethnicities, accents, and ages. \\
\textbf{VoxPopuli Accented} & Wild & A comprehensive multilingual speech corpus derived from European Parliament event recordings, featuring a rich collection of named entities. \\
\textbf{ATCOsim} & Wild & A specialized database containing ten hours of English speech from ten non-native speakers, recorded during real-time air traffic control simulations. \\ \bottomrule
\end{tabular}
}
\caption{Overview of various ASR along with brief description.}
\label{tab:list_of_datasets}
\end{table*}

%% file: tables/models.tex
\begin{table*}[h!]
\renewcommand{\arraystretch}{1}
\centering
\resizebox{\linewidth}{!}{%
\small
\begin{tabular}{p{0.45\textwidth}p{0.6\textwidth}}
\toprule
\textbf{Model Type and Models} & \textbf{Description} \\ \midrule

\textbf{nemo\_asr} \newline 
-- parakeet-ctc-1.1b \newline 
-- parakeet-ctc-0.6b \newline 
-- stt\_en\_conformer\_ctc\_large \newline 
-- stt\_en\_fastconformer\_ctc\_large \newline 
-- stt\_en\_conformer\_ctc\_small \newline 
-- parakeet-tdt-1.1b \newline 
-- parakeet-rnnt-1.1b \newline 
-- parakeet-rnnt-0.6b \newline 
-- stt\_en\_fastconformer\_transducer\_large \newline 
-- parakeet-tdt\_ctc-110m \newline 
-- canary-1b & 
NVIDIA's NeMo ASR models offer diverse architectures for speech-to-text applications. The Conformer-CTC model combines self-attention and convolutional operations, using Connectionist Temporal Classification (CTC) loss for efficient transcription. The Conformer-Transducer extends this by incorporating a Recurrent Neural Network Transducer (RNNT) decoder for autoregressive modeling. The Conformer-HAT variant separates label and blank score predictions, enhancing integration with external language models. For improved performance, the Fast-Conformer introduces depthwise convolutional subsampling, achieving approximately 2.4x faster encoding with minimal accuracy loss. \\ \hline

\textbf{speechbrain} \newline 
-- asr-wav2vec2-librispeech & 
SpeechBrain provides robust models for ASR and speaker recognition. \\ \hline

\textbf{data2vec} \newline 
-- data2vec-audio-large-960h \newline 
-- data2vec-audio-base-960h & 
Data2Vec models by Facebook are designed for speech representation learning and ASR. These models use a unified learning framework for multiple modalities. \\ \hline

\textbf{wav2vec2} \newline 
-- wav2vec2-large-960h-lv60-self \newline 
-- wav2vec2-large-robust-ft-libri-960h \newline 
-- wav2vec2-large-960h \newline 
-- wav2vec2-base-960h \newline 
-- wav2vec2-conformer-rope-large-960h-ft \newline 
-- wav2vec2-conformer-rel-pos-large-960h-ft & 
Wav2Vec2 models leverage self-supervised learning on raw audio for ASR. With advanced configurations, these models provide high accuracy for diverse speech-to-text tasks. \\ \hline

\textbf{mms} \newline 
-- mms-1b-all \newline 
-- mms-1b-fl102 & 
The Multilingual Speech (MMS) models by Facebook excel at speech recognition for multiple languages and accents. \\ \hline

\textbf{hubert} \newline 
-- hubert-xlarge-ls960-ft \newline 
-- hubert-large-ls960-ft & 
HuBERT models provide high-quality speech representations for ASR and other downstream speech tasks. \\ \hline

\textbf{seamless} \newline 
-- hf-seamless-m4t-large \newline 
-- hf-seamless-m4t-medium \newline 
-- seamless-m4t-v2-large & 
Seamless models focus on multilingual transcription and translation, offering robust real-time speech processing solutions. \\ \hline

\textbf{speechllm} \newline 
-- speechllm-1.5B \newline 
-- speechllm-2B & 
SpeechLLM models are fine-tuned for ASR and text generation, leveraging billions of parameters for high performance. \\ \hline

\textbf{whisper} \newline 
-- whisper-large-v3 \newline 
-- distil-large-v3 \newline 
-- whisper-large-v2 \newline 
-- whisper-large-v3-turbo \newline 
-- distil-large-v2 \newline 
-- whisper-large \newline 
-- whisper-tiny \newline 
-- whisper-medium.en \newline 
-- distil-small.en \newline 
-- whisper-small.en & 
Whisper models by OpenAI provide state-of-the-art transcription and translation capabilities for multilingual ASR. These models range from tiny to large configurations. \\ \hline

\textbf{moonshine} \newline 
-- moonshine-base \newline 
-- moonshine-tiny & 
Moonshine models are lightweight and optimized for efficient ASR on edge devices with minimal computational resources. \\ \bottomrule

\end{tabular}
}
\caption{Overview of various ASR along with brief description.}

\label{tab:list_of_models}
\end{table*}

%% file: tables/hyperparameters.tex
\begin{table*}[ht]
\renewcommand{\arraystretch}{1}
\centering
\resizebox{\linewidth}{!}{%
    \label{tab:hyperparameter-grid}
    \begin{tabular}{lll}
        \hline
        \textbf{Model} & \textbf{Hyperparameter} & \textbf{Values} \\
        \hline
        \multirow{4}{*}{Random Forest (RF)} 
            & \texttt{n\_estimators} & \{100, 200, 300, 500, 700, 1000\} \\
            & \texttt{max\_depth} & \{5, 10, 15, 20, 25, 30\} \\
            & \texttt{min\_samples\_split} & \{2, 5, 10, 15, 20\} \\
            & \texttt{min\_samples\_leaf} & \{1, 2, 4, 8\} \\
        \hline
        \multirow{4}{*}{Gradient Boosting (GBR)} 
            & \texttt{n\_estimators} & \{100, 200, 400, 600, 800\} \\
            & \texttt{learning\_rate} & \{0.001, 0.01, 0.05, 0.1, 0.2\} \\
            & \texttt{max\_depth} & \{3, 5, 7, 10\} \\
            & \texttt{min\_impurity\_decrease} & \{0.0, 0.001, 0.01, 0.1, 0.2\} \\
        \hline
        \multirow{4}{*}{HistGradientBoosting (HGB)} 
            & \texttt{max\_iter} & \{100, 200, 300, 400, 500\} \\
            & \texttt{learning\_rate} & \{0.001, 0.01, 0.05, 0.1, 0.2\} \\
            & \texttt{max\_depth} & \{3, 5, 7, 10, 15\} \\
            & \texttt{loss} & \{\texttt{Poisson}\} \\
        \hline
        \multirow{2}{*}{Ridge Regression (Final Estimator)}
            & \texttt{alpha} & \{1e-3, 1e-2, 0.1, 1, 10, 100, 1000\} \\
            & \texttt{positive} & \{True\} \\
        \hline
        \multirow{1}{*}{Pipeline}
            & \texttt{passthrough} & \{True\} \\
        \hline
    \end{tabular}
    }
    \caption{Hyperparameter details for regression model.}
    \label{tab:hyperparameters}
\end{table*}

%% file: sections/distribution_shift.tex
\subsection{Domain Divergence and Phonetic Diversity Analysis}\label{appsubsec:domain_shift}
\label{sec:divergence_analysis}

To quantify how our \textit{in-the-wild} evaluation sets differ from the LibriSpeech-clean corpus that was used to fine-tune both \textit{data2vec-audio-base-960h} (source) and \textit{wav2vec2-base-960h} (target), we measure acoustic domain divergence with \emph{Central Moment Discrepancy} (CMD) computed over SONAR speech embeddings and phonetic diversity with \emph{Total Vocabulary Overlap} (TVO) calculated on the corresponding transcripts. We report these numbers in Tables~\ref{tab:cmd} and~\ref{tab:tvo}.

\begin{table}[ht!]
\renewcommand{\arraystretch}{1}
\centering
\resizebox{\linewidth}{!}{%
\begin{tabular}{lcc}
\toprule
\textbf{Dataset} & \textbf{CMD} & \textbf{Interpretation} \\
\midrule
LibriSpeech (Noise)    & 0.062 & Low divergence \\
Primock57              & 0.329 & Significant divergence \\
VoxPopuli (Accented)   & 0.504 & High divergence \\
ATCOSIM                & 0.538 & High divergence \\
\bottomrule
\end{tabular}
}
\caption{Acoustic domain divergence between LibriSpeech-clean and each evaluation set, measured with Central Moment Discrepancy (CMD).}
\label{tab:cmd}
\end{table}

\begin{table}[ht!]
\renewcommand{\arraystretch}{1}
\centering
\resizebox{\linewidth}{!}{%
\begin{tabular}{lcc}
\toprule
\textbf{Dataset} & \textbf{TVO (\%)} & \textbf{Interpretation} \\
\midrule
LibriSpeech (Noise)    & 36.9 & High overlap \\
VoxPopuli (Accented)   & 20.3 & Low overlap \\
Primock57              & 14.3 & Low overlap \\
ATCOSIM                & 3.8  & Very low overlap \\
\bottomrule
\end{tabular}
}
\caption{Total Vocabulary Overlap (TVO) between LibriSpeech-clean and each evaluation set.}
\label{tab:tvo}
\end{table}

CMD values confirm that LibriSpeech (Noise) remains acoustically close to the source domain because only artificial background sounds are added, whereas Primock57, VoxPopuli (Accented), and ATCOSIM exhibit substantial acoustic shifts. TVO scores tell a complementary story in the lexical space: LibriSpeech (Noise) preserves more than one-third of the vocabulary, while Primock57, VoxPopuli (Accented), and especially ATCOSIM share far less with LibriSpeech-clean. Together, these metrics justify the \textit{in-the-wild} designation of our evaluation corpora and highlight the importance of robust models that generalize beyond clean, studio-quality speech.

%% file: tables/ablation_results_appendix.tex

\begin{table}[]
\centering
\small 
\renewcommand{\arraystretch}{0.9} 
\setlength{\tabcolsep}{5pt} 
\begin{tabular}{lcccc}
\specialrule{1.2pt}{0.5em}{0.2em} 
\multirow{2}{*}{Method} & \multirow{2}{*}{IID} & \multicolumn{3}{c}{OOD}                  \\

\cmidrule(lr){3-5}

                        &                      & D        & M       & D + M \\ 

\specialrule{1.2pt}{0.5em}{0.2em}  

Base                                    & 3.79\textsuperscript{0.16}          & 3.56\textsuperscript{0.06} & 3.76\textsuperscript{0.18} & 3.69\textsuperscript{0.06}  \\ \midrule
w/o S                                 & 3.83\textsuperscript{0.14}          & 3.65\textsuperscript{0.06} & 3.82\textsuperscript{0.16} & 3.73\textsuperscript{0.07}  \\ 
w/o PR                                  & 8.43\textsuperscript{0.28}          & 8.36\textsuperscript{0.08} & 8.67\textsuperscript{0.24} & 8.66\textsuperscript{0.08}  \\ \midrule
w/ MPR (n=2)                            & 3.69\textsuperscript{0.14}          & 3.57\textsuperscript{0.06} & 3.66\textsuperscript{0.17} & 3.69\textsuperscript{0.06}  \\
w/ MPR (n=3)                            & 3.62\textsuperscript{0.13}          & 3.44\textsuperscript{0.07} & 3.58\textsuperscript{0.15} & 3.56\textsuperscript{0.07}  \\
w/ MPR (n=4)                            & 3.57\textsuperscript{0.13}          & 3.40\textsuperscript{0.06} & 3.53\textsuperscript{0.13} & 3.52\textsuperscript{0.06}  \\
w/ MPR (n=5)                            & 3.49\textsuperscript{0.13}          & 3.37\textsuperscript{0.06} & 3.47\textsuperscript{0.12} & 3.49\textsuperscript{0.07}  \\ \midrule
w/ mMPR (n=3)                           & 3.61\textsuperscript{0.15}          & 3.40\textsuperscript{0.09} & 3.57\textsuperscript{0.13} & 3.51\textsuperscript{0.09}  \\
w/ mMPR (n=5)                           & 3.80\textsuperscript{0.15}          & 3.47\textsuperscript{0.03} & 3.77\textsuperscript{0.13} & 3.56\textsuperscript{0.04}  \\ \midrule
\multicolumn{1}{l}{Base (r=11.9)}    & 4.68\textsuperscript{0.17}          & 5.16\textsuperscript{0.06} & 4.64\textsuperscript{0.16} & 5.06\textsuperscript{0.05}  \\
\multicolumn{1}{l}{Base (r=14.0)}    & 4.84\textsuperscript{0.18}          & 4.88\textsuperscript{0.07} & 4.75\textsuperscript{0.17} & 4.77\textsuperscript{0.07}  \\
\multicolumn{1}{l}{Base (r=20.2)}    & 5.13\textsuperscript{0.12}          & 5.38\textsuperscript{0.07} & 5.12\textsuperscript{0.10} & 5.30\textsuperscript{0.07}  \\ 
\multicolumn{1}{l}{Base (r=23.5)}    & 5.60\textsuperscript{0.13}          & 6.12\textsuperscript{0.07} & 5.69\textsuperscript{0.21} & 6.03\textsuperscript{0.05}  \\ \midrule
\multicolumn{1}{l}{w/o S (r=11.9)} & 5.50\textsuperscript{0.21}          & 5.84\textsuperscript{0.06} & 5.55\textsuperscript{0.21} & 5.65\textsuperscript{0.05}  \\
\multicolumn{1}{l}{w/o S (r=14.0)} & 5.73\textsuperscript{0.12}          & 5.50\textsuperscript{0.05} & 5.71\textsuperscript{0.13} & 5.37\textsuperscript{0.06}  \\
\multicolumn{1}{l}{w/o S (r=20.2)} & 6.16\textsuperscript{0.18}          & 6.24\textsuperscript{0.08} & 6.13\textsuperscript{0.10} & 5.97\textsuperscript{0.09}  \\
\multicolumn{1}{l}{w/o S (r=23.5)} & 6.38\textsuperscript{0.09}          & 6.77\textsuperscript{0.08} & 6.43\textsuperscript{0.16} & 6.58\textsuperscript{0.08} \\ 

\specialrule{1.2pt}{0.5em}{0.2em} 

\end{tabular}
\caption{Mean absolute error between predicted character error count and actual character error count (in absolute terms) across different configurations. R - Regression, C - Classification, PR - Proxy Reference, S - Silarity, MPR - Multiple PR. The OOD results are averaged across five wild datasets. Superscript represents the standard deviation across five runs.}
\label{tab:ablation_baseline_results_cer}
\end{table}

%% file: tables/results_in_the_wild_cer.tex
\begin{table*}[ht]
\centering
\begin{tabular}{lcccc}
\toprule
Model                                     & LS\_Noise & Primock57 & Atcosim & VP\_accented \\

\midrule
asr-wav2vec2-librispeech                  & 4.2/5.8             & 17.2/20.8           & 18.8/21.9         & 9.8/14.0               \\
canary-1b                                 & 1.5/3.8             & 10.1/9.7            & 16.4/19.4         & 15.6/9.0               \\
data2vec-audio-base-960h                  & 7.0/8.1             & 20.5/23.7           & 29.5/32.0         & 13.3/17.8              \\
data2vec-audio-large-960h                 & 3.1/4.2             & 14.1/17.4           & 20.0/23.8         & 10.6/14.4              \\
distil-large-v2                           & 3.5/5.2             & 11.5/9.2            & 49.5/41.8         & 10.2/9.4               \\
distil-large-v3                           & 2.7/4.6             & 11.9/9.1            & 49.4/40.5         & 10.1/9.0               \\
distil-small.en                           & 4.2/5.8             & 12.2/10.4           & 50.7/41.8         & 9.7/9.2                \\
hf-seamless-m4t-large                     & 6.5/7.5             & 32.1/30.6           & 54.7/57.2         & 21.8/15.8              \\
hf-seamless-m4t-medium                    & 9.4/10.1            & 34.4/32.7           & 35.5/37.9         & 23.1/17.9              \\
hubert-large-ls960-ft                     & 3.0/4.2             & 14.4/17.4           & 21.3/25.0         & 10.0/14.3              \\
hubert-xlarge-ls960-ft                    & 2.7/4.1             & 15.3/18.1           & 20.1/23.8         & 10.2/14.5              \\
mms-1b-all                                & 3.6/4.8             & 19.5/19.1           & 27.2/31.8         & 17.0/12.6              \\
mms-1b-fl102                              & 9.0/10.0            & 35.0/33.2           & 55.4/57.3         & 18.2/17.6              \\
moonshine-base                            & 5.7/6.8             & 12.4/12.1           & 42.6/39.5         & 10.9/12.6              \\
moonshine-tiny                            & 8.5/9.9             & 17.9/19.0           & 38.2/38.4         & 13.2/15.1              \\
parakeet-ctc-0.6b                         & 1.7/3.7             & 10.1/9.9            & 16.2/22.7         & 9.7/9.0                \\
parakeet-ctc-1.1b                         & 1.7/3.6             & 10.0/10.1           & 14.8/21.4         & 10.0/8.0               \\
parakeet-rnnt-0.6b                        & 1.3/3.4             & 10.1/9.4            & 16.9/24.1         & 10.9/8.8               \\
parakeet-rnnt-1.1b                        & 1.3/3.3             & 9.1/9.7             & 14.5/21.3         & 11.2/7.2               \\
parakeet-tdt-1.1b                         & 1.1/3.1             & 8.2/9.4             & 14.0/20.0         & 10.9/6.8               \\
parakeet-tdt\_ctc-110m                    & 2.5/4.7             & 10.3/9.2            & 22.3/24.2         & 12.4/8.6               \\
seamless-m4t-v2-large                     & 3.5/4.6             & 24.6/23.7           & 31.6/35.8         & 25.2/19.8              \\
speechllm-1.5B                            & 9.9/11.2            & 30.1/31.4           & 85.4/88.7         & 47.3/49.0              \\
speechllm-2B                              & 8.4/9.3             & 25.3/27.7           & 33.5/36.1         & 24.0/28.3              \\
stt\_en\_conformer\_ctc\_large            & 2.1/3.4             & 8.8/11.2            & 17.1/18.2         & 11.1/7.5               \\
stt\_en\_conformer\_ctc\_small            & 4.3/5.7             & 12.7/15.6           & 21.6/23.6         & 9.5/9.7                \\
stt\_en\_fastconformer\_ctc\_large        & 3.0/5.6             & 10.1/16.3           & 17.3/25.1         & 11.5/9.2               \\
stt\_en\_fastconformer\_transducer\_large & 2.8/5.0             & 10.6/14.3           & 18.7/25.3         & 14.2/11.9              \\
wav2vec2-base-960h                        & 7.9/9.1             & 23.3/26.7           & 30.3/33.2         & 13.7/18.9              \\
wav2vec2-conformer-rel-pos-large-960h-ft  & 2.6/3.8             & 14.7/17.4           & 21.0/24.5         & 11.2/14.7              \\
wav2vec2-conformer-rope-large-960h-ft     & 2.9/4.0             & 16.1/18.7           & 22.2/25.9         & 11.0/14.2              \\
wav2vec2-large-960h                       & 5.1/6.3             & 19.1/22.4           & 28.8/31.8         & 12.2/17.4              \\
wav2vec2-large-960h-lv60-self             & 3.5/5.0             & 17.6/21.0           & 18.6/23.0         & 9.3/13.6               \\
wav2vec2-large-robust-ft-libri-960h       & 4.5/5.8             & 15.7/19.0           & 20.7/24.2         & 10.0/14.7              \\
whisper-large                             & 2.9/4.2             & 13.7/10.6           & 49.3/47.5         & 13.7/11.9              \\
whisper-large-v2                          & 2.6/3.8             & 15.3/12.5           & 48.6/51.5         & 15.3/14.2              \\
whisper-large-v3                          & 2.0/3.3             & 12.3/8.7            & 48.9/48.3         & 14.3/13.6              \\
whisper-large-v3-turbo                    & 2.0/3.2             & 12.4/8.8            & 48.3/49.9         & 19.7/19.0              \\
whisper-medium.en                         & 3.3/4.3             & 13.1/10.5           & 49.1/49.2         & 23.8/20.6              \\
whisper-small.en                          & 4.2/5.3             & 13.1/10.8           & 48.4/51.2         & 12.5/12.7              \\
whisper-tiny                              & 9.8/11.3            & 19.3/18.2           & 60.8/63.0         & 21.0/22.3      \\

\bottomrule
\end{tabular}
\caption{Actual and approximated CER ($\downarrow$), separated by a slash, on out-of-distribution wild datasets. The regression model is trained independently for each ASR model on standard benchmarks, making the wild datasets out-of-distribution.}
\label{tab:main_results_in_the_wild_cer}
\end{table*}

%% file: sections/finegrained_metrics.tex
\paragraph{Fine-grained ASR Evaluation Metrics}  
To assess whether our framework can also approximate fine-grained ASR evaluation metrics, we extend our experiments to cover insertions, deletions, and substitutions. This addition aims to evaluate the generalizability of our approach beyond overall word error rates and provide a more detailed analysis of ASR performance. We train the regression model under various configurations and examine its performance across both in-distribution (IID) and out-of-distribution (OOD) conditions. We report the results in Table~\ref{tab:finegrained_metrics}.

\input{tables/finegrained_metrics}

The results demonstrate that our regression model reliably approximates these fine-grained error counts. Specifically, the absolute differences remain low across substitutions, insertions, and deletions, even when evaluated on challenging out-of-distribution data or using transcriptions from models not seen during training. Furthermore, the similarity-only variant, which does not rely on proxy information during testing, remains competitive, highlighting the robustness and generalizability of the learned acoustic and semantic representations.

%% file: tables/finegrained_metrics.tex
\begin{table}[h]
\centering
\small
\renewcommand{\arraystretch}{0.9}
\setlength{\tabcolsep}{5pt}
\begin{tabular}{clcccc}
\specialrule{1.2pt}{0.5em}{0.2em} 

& \multirow{2}{*}{Method} & \multirow{2}{*}{IID} & \multicolumn{3}{c}{OOD}                  \\

\cmidrule(lr){4-6}

                        &           &           & D        & M       & D + M \\ 
                        
\specialrule{1.2pt}{0.5em}{0.2em} 

\multirow{7}{*}{\rotatebox{90}{Substitution}}
 & Base              & 0.76 & 0.77 & 0.77 & 0.79 \\
 & w/o S             & 0.76 & 0.76 & 0.77 & 0.78 \\
 & w/o PR            & 2.22 & 2.28 & 2.26 & 2.36 \\
 & w/ MPR (n=5)      & 0.65 & 0.68 & 0.66 & 0.69 \\
 & w/ mMPR (n=5)     & 0.68 & 0.69 & 0.69 & 0.72 \\
 & Base ($r$=17.8)   & 0.97 & 1.09 & 0.98 & 1.11 \\
 & Base ($r$=51.0)   & 1.41 & 1.39 & 1.45 & 1.42 \\
\specialrule{1pt}{0.3em}{0.3em}

\multirow{7}{*}{\rotatebox{90}{Insertion}}
 & Base              & 0.59 & 0.58 & 0.61 & 0.59 \\
 & w/o S             & 0.58 & 0.56 & 0.60 & 0.57 \\
 & w/o PR            & 0.86 & 0.94 & 0.86 & 0.95 \\
 & w/ MPR (n=5)      & 0.54 & 0.55 & 0.55 & 0.57 \\
 & w/ mMPR (n=5)     & 0.56 & 0.56 & 0.58 & 0.57 \\
 & Base ($r$=17.8)   & 0.63 & 0.80 & 0.65 & 0.82 \\
 & Base ($r$=51.0)   & 0.69 & 0.76 & 0.71 & 0.79 \\
\specialrule{1pt}{0.3em}{0.3em}

\multirow{7}{*}{\rotatebox{90}{Deletion}}
 & Base              & 0.66 & 0.63 & 0.67 & 0.64 \\
 & w/o S             & 0.69 & 0.60 & 0.69 & 0.60 \\
 & w/o PR            & 1.01 & 1.09 & 1.06 & 1.15 \\
 & w/ MPR (n=5)      & 0.60 & 0.56 & 0.60 & 0.56 \\
 & w/ mMPR (n=5)     & 0.61 & 0.57 & 0.61 & 0.58 \\
 & Base ($r$=17.8)   & 0.76 & 0.82 & 0.78 & 0.85 \\
 & Base ($r$=51.0)   & 0.92 & 1.05 & 0.97 & 1.09 \\
\specialrule{1.2pt}{0.5em}{0.2em} 

\end{tabular}
\caption{Mean absolute deviation between predicted and true counts for substitutions, insertions, and deletions. Columns: IID (source in-distribution), D (source out-of-distribution), M (target in-distribution), D+M (target out-of-distribution). Lower values indicate better approximation.}
\label{tab:finegrained_metrics}
\end{table}

%% file: figures/all_models.tex
\begin{figure*}
    \centering
\includegraphics[width=1.0\linewidth]{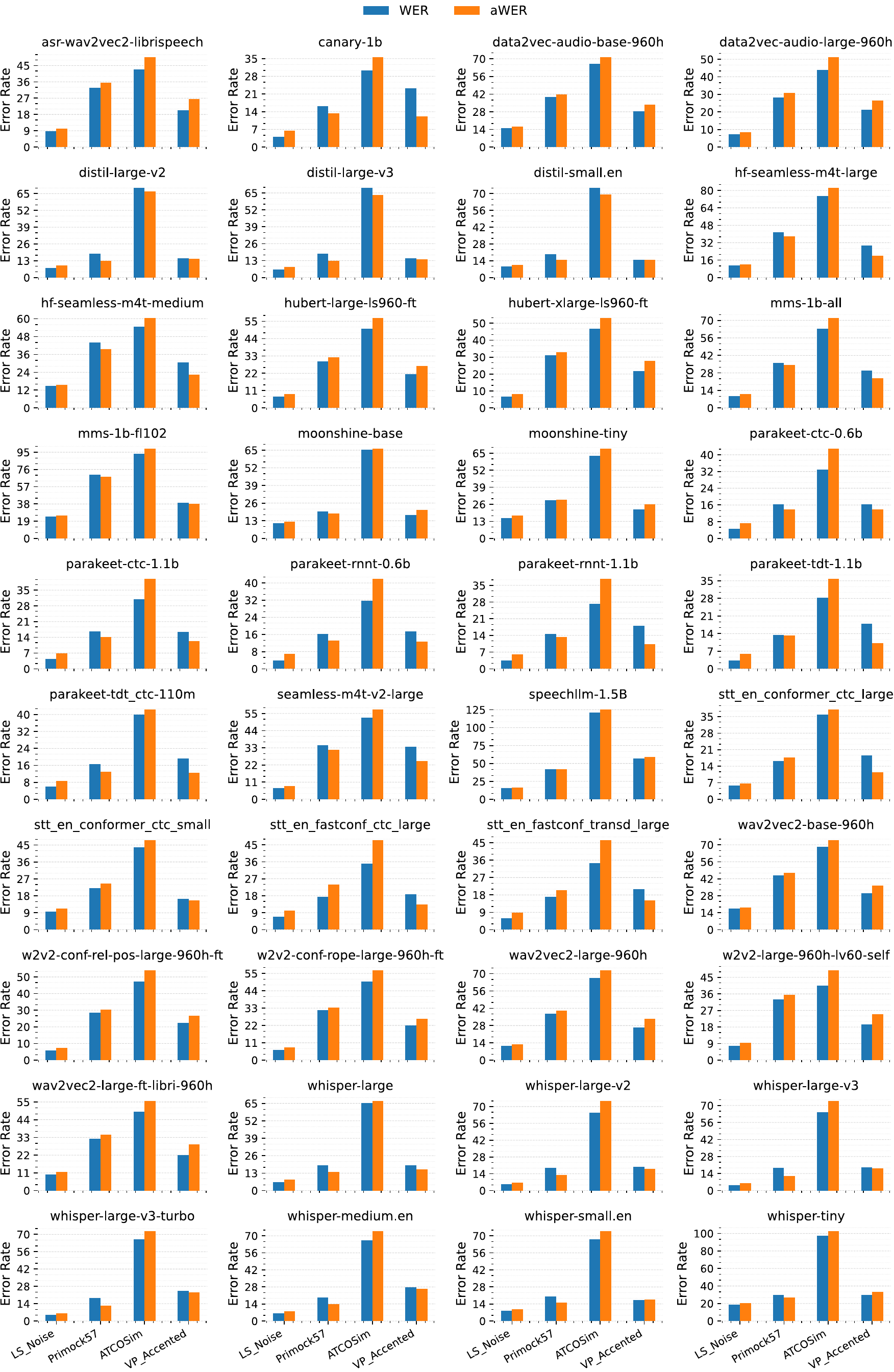}
    \caption{Actual and approximated word error rate across different models evaluated on four in-the-wild datasets.}
    \label{fig:enter-label}
\end{figure*}
\begin{figure*}
    \centering
    \includegraphics[width=1.0\linewidth]{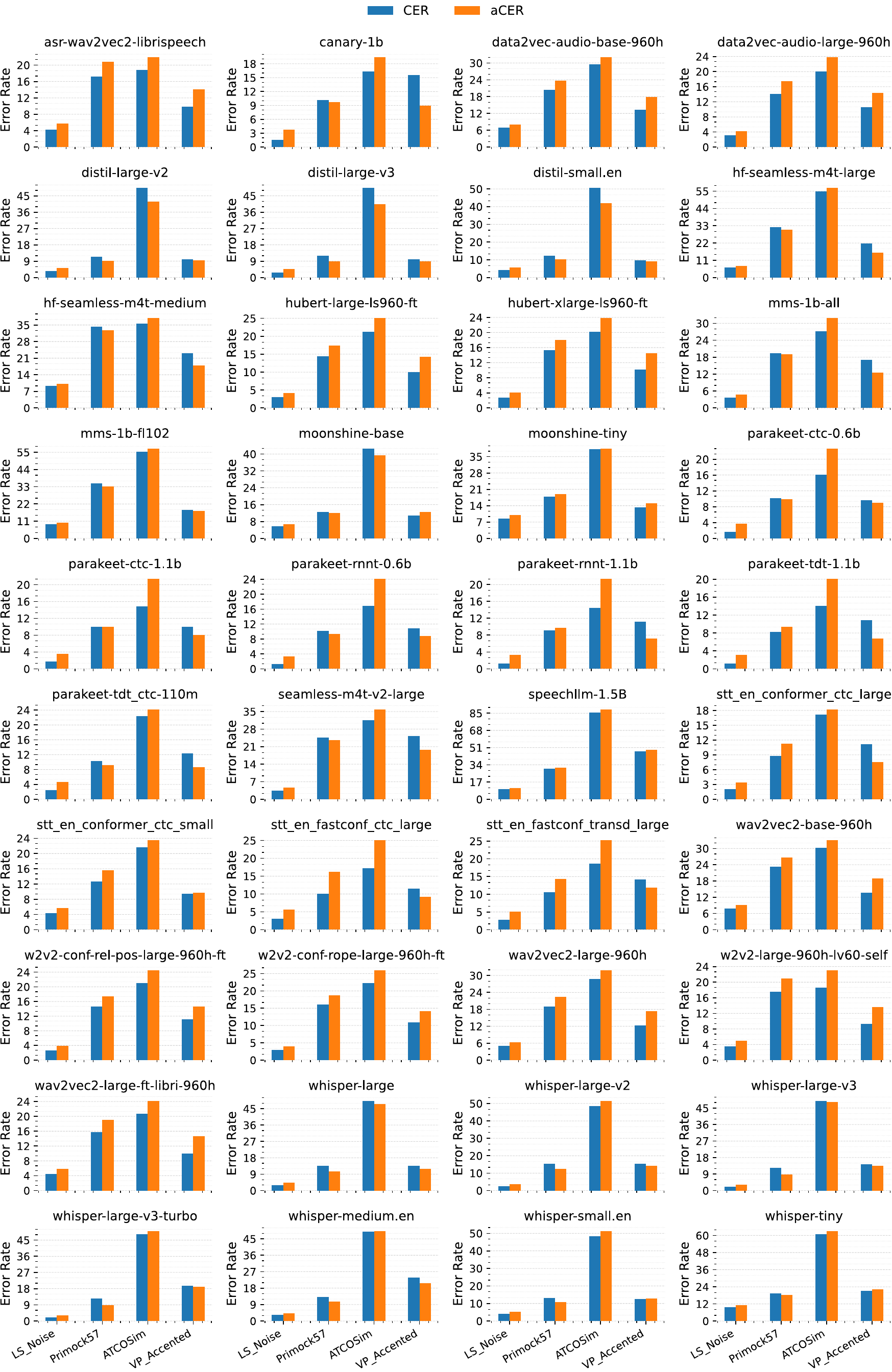}
    \caption{Actual and approximated character error rate across different models evaluated on four in-the-wild datasets.}
    \label{fig:benchmark}
\end{figure*}

%% file: tables/results_benchmark.tex
\begin{table*}[ht!]
\centering
\resizebox{\textwidth}{!}{%
\begin{tabular}{lcccccccccc}
\toprule
\multirow{2}{*}{\textbf{Model}} 
& \multicolumn{2}{c}{\textbf{AMI\_IHM}} 
& \multicolumn{2}{c}{\textbf{CV11}}
& \multicolumn{2}{c}{\textbf{Earnings22}} 
& \multicolumn{2}{c}{\textbf{Gigaspeech}} 
& \multicolumn{2}{c}{\textbf{LibriSpeech\_clean}} 
\\
\cmidrule(lr){2-3} 
\cmidrule(lr){4-5}
\cmidrule(lr){6-7}
\cmidrule(lr){8-9}
\cmidrule(lr){10-11}
& \textbf{WER/aWER} & \textbf{CER/aCER} 
& \textbf{WER/aWER} & \textbf{CER/aCER} 
& \textbf{WER/aWER} & \textbf{CER/aCER}
& \textbf{WER/aWER} & \textbf{CER/aCER}
& \textbf{WER/aWER} & \textbf{CER/aCER}
\\ 
\midrule
asr-wav2vec2-librispeech     & 28.4/30.5 & 13.8/17.6 & 25.0/29.7 & 11.7/15.0 & 37.3/33.2 & 21.3/16.1 & 16.6/16.5 & 6.9/7.4   & 1.8/3.8  & 0.5/2.2  \\
canary-1b                    & 15.4/17.6 & 9.2/12.7  & 8.7/14.2  & 4.1/8.5   & 21.8/16.0 & 15.8/9.1  & 11.1/6.9  & 5.5/4.3   & 1.5/5.7  & 0.5/3.5  \\
data2vec-audio-base-960h     & 39.9/40.4 & 19.9/23.5 & 37.8/42.3 & 18.3/21.7 & 50.8/48.6 & 28.0/25.0 & 23.8/23.5 & 10.1/10.8 & 2.8/4.0  & 0.9/1.6  \\
data2vec-audio-large-960h    & 34.1/36.1 & 16.9/21.2 & 23.3/27.9 & 10.9/14.1 & 37.7/34.5 & 21.2/16.7 & 17.0/16.6 & 7.2/7.4   & 1.8/3.9  & 0.5/1.7  \\
distil-large-v2              & 17.8/16.8 & 11.2/11.5 & 14.2/19.7 & 7.1/10.6  & 19.3/20.0 & 12.5/13.7 & 12.8/8.2  & 7.1/5.4   & 3.4/6.7  & 1.5/4.2  \\
distil-large-v3              & 18.5/17.3 & 11.6/11.7 & 13.7/19.4 & 6.6/10.3  & 18.4/19.8 & 12.1/13.0 & 12.2/7.9  & 6.9/5.3   & 2.8/6.6  & 1.2/4.1  \\
distil-small.en              & 18.5/18.4 & 11.1/12.6 & 18.5/23.1 & 9.4/12.5  & 21.2/21.4 & 13.6/14.7 & 13.1/8.6  & 7.3/5.7   & 3.7/7.6  & 1.6/4.5  \\
hf-seamless-m4t-large        & 36.3/33.9 & 25.4/25.1 & 9.5/13.2  & 5.1/7.4   & 30.7/32.8 & 21.1/23.9 & 24.2/21.1 & 16.7/15.7 & 3.2/4.8  & 1.5/2.7  \\
hf-seamless-m4t-medium       & 40.6/37.2 & 29.5/28.9 & 11.3/14.3 & 6.0/7.4   & 33.7/35.9 & 23.9/26.4 & 30.2/28.1 & 22.3/21.7 & 3.8/5.3  & 1.6/2.9  \\
hubert-large-ls960-ft        & 31.1/33.6 & 15.2/19.8 & 24.1/28.8 & 10.6/13.6 & 37.6/34.4 & 20.6/16.3 & 19.3/18.3 & 8.1/7.8   & 2.1/3.7  & 0.6/1.6  \\
hubert-xlarge-ls960-ft       & 31.1/34.3 & 15.0/20.0 & 24.1/28.7 & 10.5/13.9 & 37.3/34.9 & 20.4/15.9 & 18.1/17.4 & 7.3/7.6   & 2.0/3.8  & 0.6/1.7  \\
mms-1b-all                   & 37.0/36.2 & 19.1/20.8 & 22.5/27.5 & 8.9/12.5  & 34.1/30.6 & 19.6/15.1 & 19.4/16.9 & 8.3/7.6   & 4.2/6.2  & 1.3/2.7  \\
mms-1b-fl102                 & 75.4/73.3 & 35.1/33.9 & 42.6/45.3 & 17.8/19.9 & 50.6/52.3 & 24.2/26.5 & 37.2/35.7 & 15.7/15.2 & 15.8/17.3 & 5.1/5.9  \\
moonshine-base               & 15.6/24.7 & 9.4/16.7  & 20.8/25.4 & 10.8/13.8 & 24.3/25.6 & 15.9/16.6 & 14.2/10.4 & 8.1/6.8   & 3.4/6.3  & 1.3/3.7  \\
moonshine-tiny               & 21.3/25.3 & 12.8/16.7 & 26.7/31.7 & 14.4/17.3 & 31.2/32.7 & 19.7/20.2 & 16.6/14.1 & 9.1/8.6   & 4.5/7.2  & 1.8/4.2  \\
parakeet-ctc-0.6b            & 17.0/23.1 & 10.0/16.3 & 10.7/21.1 & 5.1/11.2  & 24.7/19.1 & 16.9/11.5 & 12.0/8.6  & 6.1/5.2   & 2.0/5.1  & 0.7/2.5  \\
parakeet-ctc-1.1b            & 15.7/21.4 & 9.0/15.3  & 10.5/20.1 & 5.2/11.0  & 24.0/17.7 & 16.6/10.7 & 12.2/7.9  & 6.2/5.0   & 1.8/5.4  & 0.5/2.6  \\
parakeet-rnnt-0.6b           & 18.8/24.0 & 11.7/17.9 & 8.5/19.9  & 4.2/10.8  & 25.2/18.7 & 17.5/11.5 & 11.7/9.0  & 6.2/5.4   & 1.8/5.5  & 0.6/3.2  \\
parakeet-rnnt-1.1b           & 18.6/23.5 & 11.7/17.2 & 6.7/19.6  & 3.4/10.5  & 25.7/17.9 & 18.4/11.4 & 11.3/8.4  & 6.0/5.0   & 1.5/5.0  & 0.5/3.3  \\
parakeet-tdt-1.1b            & 17.1/23.5 & 10.2/16.9 & 7.2/19.6  & 3.4/10.6  & 24.5/16.6 & 17.1/10.0 & 10.2/7.8  & 4.9/4.7   & 1.3/6.0  & 0.4/2.9  \\
parakeet-tdt\_ctc-110m       & 18.5/18.8 & 10.7/13.6 & 12.7/17.7 & 6.9/10.1  & 22.2/14.8 & 15.7/9.2  & 12.6/8.2  & 6.2/5.0   & 2.6/6.7  & 0.9/3.8  \\
seamless-m4t-v2-large        & 43.0/42.3 & 30.2/30.2 & 8.2/12.3  & 3.9/6.3   & 47.3/47.4 & 33.7/33.9 & 25.7/23.2 & 18.1/17.2 & 2.7/4.4  & 1.0/2.5  \\
speechllm-1.5B               & 67.7/69.3 & 51.5/55.0 & 18.5/22.7 & 10.0/12.7 & 50.8/48.2 & 38.3/35.4 & 27.5/26.0 & 18.1/18.3 & 10.5/12.1 & 7.3/9.2  \\
speechllm-2B                 & 38.6/40.8 & 24.3/28.2 & 24.6/28.2 & 16.5/18.3 & 47.3/45.0 & 32.5/30.8 & 24.4/23.6 & 13.5/13.8 & 7.0/9.3  & 4.5/4.8  \\
stt\_en\_conformer\_ctc\_large & 15.3/19.9 & 7.9/13.4  & 10.4/15.4 & 4.7/8.0   & 24.8/20.0 & 16.4/10.7 & 13.2/10.6 & 5.9/5.6   & 2.2/3.7  & 0.7/2.4  \\
stt\_en\_conformer\_ctc\_small & 21.2/24.4 & 11.2/15.4 & 19.1/24.1 & 8.9/12.2  & 29.3/25.3 & 19.0/14.1 & 15.5/14.9 & 7.2/7.7   & 3.9/5.4  & 1.4/3.1  \\
stt\_en\_fastconformer\_ctc\_large & 20.3/24.0 & 11.7/15.3 & 9.5/19.3  & 4.6/10.2  & 27.3/21.5 & 18.3/13.0 & 14.5/14.7 & 7.2/8.2   & 1.9/5.2  & 0.7/2.8  \\
stt\_en\_fastconformer\_transducer\_large 
& 19.8/22.0 & 12.9/16.8 & 9.3/18.0  & 4.7/9.8   & 31.5/26.9 & 23.0/18.8 & 13.6/13.2 & 7.4/7.8   & 1.8/3.9  & 0.6/2.6  \\
wav2vec2-base-960h           & 37.9/38.7 & 18.7/21.9 & 40.6/45.7 & 19.5/22.8 & 51.1/48.6 & 28.2/25.4 & 26.2/26.6 & 11.7/12.2 & 3.7/4.5  & 1.1/1.9  \\
wav2vec2-conformer-rel-pos-large-960h-ft 
& 35.0/38.7 & 18.5/24.1 & 23.7/28.0 & 10.7/13.7 & 38.4/36.2 & 21.7/17.6 & 18.5/17.2 & 8.5/7.9   & 1.6/3.3  & 0.5/1.5  \\
wav2vec2-conformer-rope-large-960h-ft
& 34.3/36.4 & 18.0/22.7 & 23.6/28.5 & 11.6/15.0 & 36.9/33.9 & 21.4/16.7 & 17.9/17.7 & 7.3/7.6   & 1.8/3.8  & 0.5/1.6  \\
wav2vec2-large-960h          & 34.0/36.4 & 16.4/20.2 & 34.1/38.6 & 16.2/19.4 & 46.4/43.4 & 25.4/21.7 & 20.6/20.5 & 8.6/9.1   & 2.9/4.3  & 0.8/2.1  \\
wav2vec2-large-960h-lv60-self
& 29.1/31.5 & 15.5/19.5 & 23.1/28.8 & 11.0/15.0 & 36.7/32.5 & 20.8/15.7 & 17.6/17.2 & 7.5/8.0   & 1.7/3.5  & 0.5/1.9  \\
wav2vec2-large-robust-ft-libri-960h
& 30.5/33.9 & 13.8/19.2 & 25.0/29.3 & 10.7/13.8 & 37.1/33.5 & 20.5/15.7 & 18.0/17.6 & 7.1/7.7   & 2.8/4.3  & 0.8/2.3  \\
whisper-large                 & 18.5/18.3 & 12.3/13.0 & 13.0/18.0 & 6.6/9.3   & 18.8/20.3 & 12.3/14.9 & 12.2/7.7  & 7.1/5.1   & 2.8/5.1  & 1.4/3.5  \\
whisper-large-v2              & 18.6/17.1 & 12.1/11.8 & 11.3/15.5 & 5.7/8.0   & 19.0/21.5 & 13.0/15.8 & 12.5/7.1  & 7.3/4.9   & 2.8/5.1  & 1.5/3.2  \\
whisper-large-v3              & 19.0/17.1 & 12.3/12.0 & 9.9/14.5  & 4.9/6.9   & 18.2/20.7 & 12.1/14.9 & 12.5/7.3  & 7.2/4.9   & 2.2/4.0  & 0.9/2.9  \\
whisper-large-v3-turbo        & 19.0/17.5 & 12.3/11.9 & 12.6/16.1 & 6.3/8.1   & 18.8/21.1 & 12.9/15.6 & 12.2/6.8  & 7.1/4.6   & 2.4/4.4  & 1.1/2.5  \\
whisper-medium.en             & 20.3/18.8 & 13.7/14.5 & 14.3/17.8 & 7.2/9.2   & 20.1/22.6 & 13.3/16.5 & 12.8/7.6  & 7.6/5.5   & 3.3/5.5  & 1.8/3.5  \\
whisper-small.en              & 19.8/17.9 & 12.8/12.2 & 17.8/21.9 & 9.3/11.3  & 20.6/22.9 & 13.6/16.3 & 12.8/8.1  & 7.3/5.1   & 3.3/5.6  & 1.4/3.1  \\
whisper-tiny                  & 26.7/25.1 & 16.7/17.0 & 33.5/40.3 & 17.7/20.9 & 33.8/35.4 & 22.0/25.4 & 20.6/18.2 & 12.0/11.0 & 7.9/11.2 & 3.4/5.1  \\
\bottomrule
\end{tabular}
}
\caption{Actual and approximated WER and CER, separated by a slash, across five standard datasets. The regression model is trained on nine datasets and tested on one, with this process repeated for all datasets, ensuring that the test data is always out-of-distribution.}
\label{tab:results_benchmark_part1}
\end{table*}

\begin{table*}[ht!]
\centering
\resizebox{\textwidth}{!}{%
\begin{tabular}{lcccccccccc}
\toprule
\multirow{2}{*}{\textbf{Model}} 
& \multicolumn{2}{c}{\textbf{peoples\_speech}} 
& \multicolumn{2}{c}{\textbf{slue\_voxceleb}}
& \multicolumn{2}{c}{\textbf{spgispeech\_S}} 
& \multicolumn{2}{c}{\textbf{tedlium-dev-test}}
& \multicolumn{2}{c}{\textbf{voxpopuli\_en}} 
\\
\cmidrule(lr){2-3} 
\cmidrule(lr){4-5}
\cmidrule(lr){6-7}
\cmidrule(lr){8-9}
\cmidrule(lr){10-11}
& \textbf{WER/aWER} & \textbf{CER/aCER} 
& \textbf{WER/aWER} & \textbf{CER/aCER} 
& \textbf{WER/aWER} & \textbf{CER/aCER}
& \textbf{WER/aWER} & \textbf{CER/aCER}
& \textbf{WER/aWER} & \textbf{CER/aCER}
\\
\midrule
asr-wav2vec2-librispeech & 
35.6/32.9 & 19.8/17.7 & 19.5/20.4 & 9.8/12.2 & 11.1/12.2 & 4.8/4.7 & 10.3/11.1 & 5.2/5.8 & 14.3/12.6 & 6.6/5.1 \\

canary-1b & 
16.5/22.5 & 11.1/15.2 & 14.9/11.1 & 10.8/8.2 & 3.2/6.7 & 2.0/3.9 & 7.9/7.6 & 5.9/5.0 & 6.4/4.9 & 3.9/3.4 \\

data2vec-audio-base-960h & 
43.4/38.6 & 24.4/20.8 & 26.1/27.6 & 13.0/15.5 & 19.2/19.8 & 8.2/7.9 & 13.6/14.2 & 6.3/6.4 & 18.9/17.5 & 8.5/7.1 \\

data2vec-audio-large-960h & 
35.1/31.3 & 20.0/17.3 & 20.4/22.1 & 10.3/12.9 & 11.3/12.0 & 4.9/4.7 & 9.9/10.6 & 4.5/5.0 & 14.9/13.4 & 6.9/5.5 \\

distil-large-v2 & 
17.4/21.8 & 12.2/14.1 & 16.0/10.8 & 11.4/7.4 & 3.7/7.6 & 1.8/4.5 & 10.4/8.5 & 8.8/5.4 & 9.5/8.2 & 5.8/4.6 \\

distil-large-v3 & 
17.4/21.6 & 12.4/13.8 & 14.4/10.0 & 10.3/6.8 & 3.6/7.4 & 1.8/4.5 & 10.7/9.2 & 8.6/5.7 & 9.3/6.7 & 5.8/4.1 \\

distil-small.en & 
19.0/22.5 & 13.3/14.3 & 15.9/11.4 & 11.3/7.8 & 4.0/7.9 & 1.9/4.7 & 10.8/8.8 & 9.1/5.6 & 10.2/7.4 & 6.4/4.3 \\

hf-seamless-m4t-large & 
38.5/41.5 & 29.2/30.1 & 47.2/42.8 & 39.4/36.1 & 16.2/18.7 & 11.5/13.0 & 19.8/19.1 & 15.7/14.4 & 8.1/6.5 & 5.0/3.6 \\

hf-seamless-m4t-medium & 
43.6/45.7 & 33.6/34.2 & 50.9/47.4 & 43.2/40.3 & 12.9/15.5 & 8.8/10.4 & 27.0/26.2 & 21.3/20.0 & 8.8/7.3 & 5.5/4.4 \\

hubert-large-ls960-ft & 
34.1/31.3 & 18.8/17.7 & 20.8/22.0 & 10.1/12.3 & 11.6/12.4 & 4.9/4.6 & 11.0/12.0 & 5.3/5.4 & 15.0/13.6 & 6.9/5.4 \\

hubert-xlarge-ls960-ft & 
35.5/31.5 & 19.5/16.0 & 20.3/22.1 & 9.9/12.3  & 11.9/12.3 & 4.8/4.5 & 10.1/11.2 & 4.2/5.0 & 14.5/12.8 & 6.7/5.3 \\

mms-1b-all & 
32.2/36.0 & 16.8/18.7 & 27.6/26.3 & 14.6/14.8 & 10.0/12.5 & 3.8/4.9 & 13.5/13.3 & 7.3/6.5 & 8.9/7.6 & 4.4/3.1 \\

mms-1b-fl102 & 
52.4/52.7 & 26.0/25.5 & 51.7/48.7 & 26.1/23.2 & 19.1/22.9 & 5.9/8.6 & 29.7/30.3 & 12.9/12.9 & 22.6/20.5 & 9.3/7.9 \\

moonshine-base & 
26.4/26.2 & 18.1/17.5 & 17.0/13.8 & 11.6/9.1 & 6.4/7.5 & 3.4/3.9 & 5.8/7.0 & 3.4/4.1 & 11.7/9.9 & 6.7/4.6 \\

moonshine-tiny & 
31.8/30.8 & 20.5/19.1 & 20.1/17.2 & 13.4/11.8 & 9.1/9.9 & 4.8/5.3 & 9.8/9.5 & 6.7/5.8 & 14.9/12.9 & 8.2/7.2 \\

parakeet-ctc-0.6b & 
24.2/20.0 & 16.5/12.6 & 13.1/11.0 & 8.7/7.8  & 6.4/7.5 & 3.6/3.8 & 4.3/7.2 & 2.6/4.4 & 7.0/7.2 & 4.1/3.7 \\

parakeet-ctc-1.1b & 
20.7/18.1 & 13.9/11.8 & 13.0/11.6 & 8.7/8.3  & 6.4/7.1 & 3.7/3.6 & 5.0/7.4 & 3.0/4.4 & 6.7/6.5 & 3.9/3.3 \\

parakeet-rnnt-0.6b & 
21.9/17.9 & 15.4/12.2 & 14.5/11.6 & 10.0/8.6 & 4.9/7.3 & 2.9/3.7 & 5.0/6.9 & 3.0/4.0 & 6.4/6.7 & 3.8/3.7 \\

parakeet-rnnt-1.1b & 
23.3/17.2 & 16.6/11.8 & 14.1/10.9 & 9.8/8.9  & 4.5/7.7 & 2.7/4.4 & 5.2/7.7 & 3.5/4.9 & 5.6/6.0 & 3.4/3.2 \\

parakeet-tdt-1.1b & 
24.5/17.9 & 16.6/12.6 & 13.5/10.6 & 9.1/7.9  & 5.4/7.7 & 3.2/4.2 & 4.4/7.1 & 2.8/4.4 & 5.5/6.0 & 3.3/3.1 \\

parakeet-tdt\_ctc-110m & 
16.6/21.4 & 11.5/15.1 & 15.3/11.5 & 10.7/8.3 & 3.8/7.3 & 2.2/4.0 & 5.2/6.6 & 3.3/4.1 & 7.5/6.2 & 4.6/3.1 \\

seamless-m4t-v2-large & 
35.0/36.3 & 25.1/24.9 & 45.1/43.2 & 35.9/34.7 & 11.7/13.6 & 7.6/8.7 & 26.5/25.5 & 21.0/19.1 & 8.0/7.8 & 5.7/5.0 \\

speechllm-1.5B & 
45.1/44.5 & 32.7/32.0 & 60.3/60.8 & 44.1/46.8 & 10.6/10.9 & 6.2/5.8 & 19.4/17.6 & 14.2/11.9 & 30.8/29.9 & 22.0/21.3 \\

speechllm-2B & 
52.9/53.1 & 36.6/35.7 & 36.9/37.8 & 25.8/27.3 & 14.6/15.3 & 8.1/7.5 & 18.8/16.7 & 12.9/9.4 & 28.7/27.7 & 18.5/17.5 \\

stt\_en\_conformer\_ctc\_large & 
24.2/21.5 & 15.2/13.0 & 12.6/14.7 & 7.6/9.4  & 7.9/7.3 & 4.1/3.4 & 5.9/7.7 & 3.3/4.4 & 6.9/5.3 & 3.9/2.7 \\

stt\_en\_conformer\_ctc\_small & 
31.3/26.9 & 19.0/15.7 & 16.6/17.8 & 9.6/11.7 & 10.0/9.4 & 5.1/4.1 & 8.0/9.9 & 3.9/5.2 & 8.9/7.3 & 5.0/3.8 \\

stt\_en\_fastconformer\_ctc\_large & 
26.9/20.5 & 18.3/12.9 & 15.4/14.0 & 9.9/9.7  & 6.9/8.4 & 3.7/4.1 & 5.7/7.8 & 3.1/4.5 & 6.3/6.0 & 3.8/3.3 \\

stt\_en\_fastconformer\_transducer\_large & 
26.5/20.4 & 19.0/13.8 & 16.9/15.1 & 11.3/11.1 & 6.0/7.9 & 3.4/4.0 & 4.9/7.0 & 2.8/4.4 & 6.7/6.9 & 4.1/3.8 \\

wav2vec2-base-960h & 
44.7/40.1 & 24.5/20.6 & 27.3/28.7 & 13.5/15.7 & 21.5/22.4 & 8.9/8.7 & 13.8/14.7 & 6.1/6.6 & 20.5/19.4 & 9.1/7.8 \\

wav2vec2-conformer-rel-pos-large-960h-ft & 
37.3/34.7 & 21.2/19.7 & 20.3/22.1 & 10.6/13.4 & 12.0/12.2 & 5.2/4.6 & 11.7/12.3 & 6.6/6.3 & 14.8/13.1 & 6.9/5.5 \\

wav2vec2-conformer-rope-large-960h-ft & 
35.3/32.1 & 20.4/19.3 & 20.6/22.1 & 10.4/12.9 & 11.7/12.5 & 5.1/4.8 & 10.9/11.8 & 5.5/6.0 & 14.5/13.4 & 6.9/5.4 \\

wav2vec2-large-960h & 
38.9/35.7 & 21.6/19.2 & 23.2/25.1 & 11.5/13.7 & 16.3/17.1 & 6.9/6.6 & 12.2/13.2 & 5.6/6.1 & 18.1/16.7 & 8.2/7.0 \\

wav2vec2-large-960h-lv60-self & 
32.5/29.4 & 18.7/15.7 & 20.2/20.8 & 10.7/12.9 & 10.4/12.2 & 4.3/4.7 & 9.5/10.5 & 4.2/4.9 & 13.5/12.5 & 6.4/5.2 \\

wav2vec2-large-robust-ft-libri-960h & 
36.2/32.6 & 19.2/16.9 & 20.9/23.0 & 9.6/12.5  & 11.8/12.5 & 5.0/4.8 & 10.6/11.9 & 4.8/5.4 & 15.4/14.0 & 7.0/5.8 \\

whisper-large & 
31.2/32.4 & 24.6/23.0 & 17.6/12.6 & 13.7/10.5 & 3.7/7.4 & 2.1/4.4 & 19.3/16.1 & 14.0/10.3 & 8.9/6.0 & 5.5/3.2 \\

whisper-large-v2 & 
18.8/25.2 & 14.2/18.1 & 18.7/15.1 & 14.8/12.1 & 4.1/7.7 & 2.4/4.8 & 28.3/25.3 & 19.4/14.4 & 8.7/6.8 & 5.5/3.8 \\

whisper-large-v3 & 
20.4/27.5 & 15.6/19.5 & 15.6/11.9 & 11.8/8.7 & 3.2/6.3 & 1.7/4.0 & 10.5/8.3 & 8.7/5.5 & 11.0/8.6 & 7.8/6.1 \\

whisper-large-v3-turbo & 
16.0/23.7 & 12.0/16.5 & 15.3/11.9 & 11.5/9.0 & 3.1/6.3 & 1.7/3.9 & 9.9/8.1 & 8.5/5.2 & 13.3/11.1 & 9.8/7.9 \\

whisper-medium.en & 
20.1/25.1 & 15.3/18.0 & 21.2/16.1 & 16.6/13.2 & 4.0/7.7 & 2.2/5.0 & 17.3/14.6 & 18.3/14.2 & 9.0/7.2 & 5.6/3.2 \\

whisper-small.en & 
21.2/25.1 & 16.5/18.0 & 18.2/14.1 & 13.9/10.9 & 3.9/7.5 & 2.1/4.6 & 10.6/8.3 & 15.6/12.0 & 9.5/8.1 & 5.9/3.4 \\

whisper-tiny & 
30.1/31.9 & 21.7/21.5 & 24.0/20.3 & 17.5/14.4 & 8.1/11.9 & 3.9/6.7 & 17.6/15.3 & 13.0/9.5 & 13.2/11.2 & 7.4/6.3 \\

\bottomrule
\end{tabular}
}
\caption{Actual and approximated WER and CER, separated by a forward slash, across five standard datasets. The regression model is trained on nine datasets and tested on one, with this process repeated for all datasets, ensuring that the test data is always out-of-distribution.}
\label{tab:results_benchmark_part2}
\end{table*}

%% file: acl_latex.bbl
\begin{thebibliography}{68}
\providecommand{\natexlab}[1]{#1}

\bibitem[{Ali and Renals(2018)}]{ali-renals-2018-word}
Ahmed Ali and Steve Renals. 2018.
\newblock \href {https://doi.org/10.18653/v1/P18-2004} {Word error rate estimation for speech recognition: e-{WER}}.
\newblock In \emph{Proceedings of the 56th Annual Meeting of the Association for Computational Linguistics (Volume 2: Short Papers)}, pages 20--24, Melbourne, Australia. Association for Computational Linguistics.

\bibitem[{Ali and Renals(2020)}]{Ali2020WordER}
Ahmed~M. Ali and Steve Renals. 2020.
\newblock \href {https://api.semanticscholar.org/CorpusID:221090197} {Word error rate estimation without asr output: e-wer2}.
\newblock In \emph{Interspeech}.

\bibitem[{Ao et~al.(2022)Ao, Wang, Zhou, Wang, Ren, Wu, Liu, Ko, Li, Zhang, Wei, Qian, Li, and Wei}]{ao2022speecht5unifiedmodalencoderdecoderpretraining}
Junyi Ao, Rui Wang, Long Zhou, Chengyi Wang, Shuo Ren, Yu~Wu, Shujie Liu, Tom Ko, Qing Li, Yu~Zhang, Zhihua Wei, Yao Qian, Jinyu Li, and Furu Wei. 2022.
\newblock \href {https://arxiv.org/abs/2110.07205} {Speecht5: Unified-modal encoder-decoder pre-training for spoken language processing}.
\newblock \emph{Preprint}, arXiv:2110.07205.

\bibitem[{Ardila et~al.(2020)Ardila, Branson, Davis, Henretty, Kohler, Meyer, Morais, Saunders, Tyers, and Weber}]{ardila2020common}
Rosana Ardila, Megan Branson, Kelly Davis, Michael Henretty, Michael Kohler, Josh Meyer, Reuben Morais, Lindsay Saunders, Francis~M. Tyers, and Gregor Weber. 2020.
\newblock Common voice: A massively-multilingual speech corpus.
\newblock \url{https://commonvoice.mozilla.org/en/datasets}.
\newblock Accessed: [Insert Date].

\bibitem[{Baevski et~al.(2022)Baevski, Hsu, Xu, Babu, Gu, and Auli}]{baevski2022data2vecgeneralframeworkselfsupervised}
Alexei Baevski, Wei-Ning Hsu, Qiantong Xu, Arun Babu, Jiatao Gu, and Michael Auli. 2022.
\newblock \href {https://arxiv.org/abs/2202.03555} {data2vec: A general framework for self-supervised learning in speech, vision and language}.
\newblock \emph{Preprint}, arXiv:2202.03555.

\bibitem[{Baevski et~al.(2020)Baevski, Zhou, Mohamed, and Auli}]{baevski2020wav2vec20frameworkselfsupervised}
Alexei Baevski, Henry Zhou, Abdelrahman Mohamed, and Michael Auli. 2020.
\newblock \href {https://arxiv.org/abs/2006.11477} {wav2vec 2.0: A framework for self-supervised learning of speech representations}.
\newblock \emph{Preprint}, arXiv:2006.11477.

\bibitem[{Barrault et~al.(2025)Barrault, Chung, Meglioli, Dale, Dong, Duquenne, Elsahar, Gong, Heffernan, Hoffman, Klaiber, Li, Licht, Maillard, Rakotoarison, Sadagopan, Wenzek, Ye, Akula, Chen, El~Hachem, Ellis, Gonzalez, Haaheim, Hansanti, Howes, Huang, Hwang, Inaguma, Jain, Kalbassi, Kallet, Kulikov, Lam, Li, Ma, Mavlyutov, Peloquin, Ramadan, Ramakrishnan, Sun, Tran, Tran, Tufanov, Vogeti, Wood, Yang, Yu, Andrews, Balioglu, Costa-juss{\`a}, {\c{C}}elebi, Elbayad, Gao, Guzm{\'a}n, Kao, Lee, Mourachko, Pino, Popuri, Ropers, Saleem, Schwenk, Tomasello, Wang, Wang, Wang, and Team}]{Barrault2025}
Lo{\"i}c Barrault, Yu-An Chung, Mariano~Coria Meglioli, David Dale, Ning Dong, Paul-Ambroise Duquenne, Hady Elsahar, Hongyu Gong, Kevin Heffernan, John Hoffman, Christopher Klaiber, Pengwei Li, Daniel Licht, Jean Maillard, Alice Rakotoarison, Kaushik~Ram Sadagopan, Guillaume Wenzek, Ethan Ye, Bapi Akula, Peng-Jen Chen, Naji El~Hachem, Brian Ellis, Gabriel~Mejia Gonzalez, Justin Haaheim, Prangthip Hansanti, Russ Howes, Bernie Huang, Min-Jae Hwang, Hirofumi Inaguma, Somya Jain, Elahe Kalbassi, Amanda Kallet, Ilia Kulikov, Janice Lam, Daniel Li, Xutai Ma, Ruslan Mavlyutov, Benjamin Peloquin, Mohamed Ramadan, Abinesh Ramakrishnan, Anna Sun, Kevin Tran, Tuan Tran, Igor Tufanov, Vish Vogeti, Carleigh Wood, Yilin Yang, Bokai Yu, Pierre Andrews, Can Balioglu, Marta~R. Costa-juss{\`a}, Onur {\c{C}}elebi, Maha Elbayad, Cynthia Gao, Francisco Guzm{\'a}n, Justine Kao, Ann Lee, Alexandre Mourachko, Juan Pino, Sravya Popuri, Christophe Ropers, Safiyyah Saleem, Holger Schwenk, Paden Tomasello, Changhan Wang, Jeff Wang,
  Skyler Wang, and SEAMLESS~Communication Team. 2025.
\newblock \href {https://doi.org/10.1038/s41586-024-08359-z} {Joint speech and text machine translation for up to 100 languages}.
\newblock \emph{Nature}, 637(8046):587--593.

\bibitem[{Carletta et~al.(2005)Carletta, Ashby, Bourban, Flynn, Guillemot, Hain, Kadlec, Karaiskos, Kraaij, Kronenthal et~al.}]{ami2005corpus}
Jean Carletta, Simone Ashby, Sebastien Bourban, Mike Flynn, Mael Guillemot, Thomas Hain, Jaroslav Kadlec, Vasilis Karaiskos, Wessel Kraaij, Melissa Kronenthal, et~al. 2005.
\newblock The ami meeting corpus: A pre-announcement.
\newblock \url{https://groups.inf.ed.ac.uk/ami/corpus/}.
\newblock Accessed: [Insert Date].

\bibitem[{Chen et~al.(2021)Chen, Chai, Wang, Du, Zhang, Weng, Su, Povey, Trmal, Zhang et~al.}]{chen2021gigaspeech}
Guoguo Chen, Shuzhou Chai, Guanbo Wang, Jiayu Du, Wei-Qiang Zhang, Chao Weng, Dan Su, Daniel Povey, Jan Trmal, Junbo Zhang, et~al. 2021.
\newblock Gigaspeech: An evolving, multi-domain asr corpus with 10,000 hours of transcribed audio.
\newblock \emph{arXiv preprint arXiv:2106.06909}.

\bibitem[{Chowdhury and Ali(2023)}]{chowdhury2023multilingualworderrorrate}
Shammur~Absar Chowdhury and Ahmed Ali. 2023.
\newblock \href {https://arxiv.org/abs/2304.00649} {Multilingual word error rate estimation: e-wer3}.
\newblock \emph{Preprint}, arXiv:2304.00649.

\bibitem[{Communication et~al.(2023)Communication, Barrault, Chung, Meglioli, Dale, Dong, Duquenne, Elsahar, Gong, Heffernan, Hoffman, Klaiber, Li, Licht, Maillard, Rakotoarison, Sadagopan, Wenzek, Ye, Akula, Chen, Hachem, Ellis, Gonzalez, Haaheim, Hansanti, Howes, Huang, Hwang, Inaguma, Jain, Kalbassi, Kallet, Kulikov, Lam, Li, Ma, Mavlyutov, Peloquin, Ramadan, Ramakrishnan, Sun, Tran, Tran, Tufanov, Vogeti, Wood, Yang, Yu, Andrews, Balioglu, Costa-jussà, Celebi, Elbayad, Gao, Guzmán, Kao, Lee, Mourachko, Pino, Popuri, Ropers, Saleem, Schwenk, Tomasello, Wang, Wang, and Wang}]{communication2023seamlessm4tmassivelymultilingual}
Seamless Communication, Loïc Barrault, Yu-An Chung, Mariano~Cora Meglioli, David Dale, Ning Dong, Paul-Ambroise Duquenne, Hady Elsahar, Hongyu Gong, Kevin Heffernan, John Hoffman, Christopher Klaiber, Pengwei Li, Daniel Licht, Jean Maillard, Alice Rakotoarison, Kaushik~Ram Sadagopan, Guillaume Wenzek, Ethan Ye, Bapi Akula, Peng-Jen Chen, Naji~El Hachem, Brian Ellis, Gabriel~Mejia Gonzalez, Justin Haaheim, Prangthip Hansanti, Russ Howes, Bernie Huang, Min-Jae Hwang, Hirofumi Inaguma, Somya Jain, Elahe Kalbassi, Amanda Kallet, Ilia Kulikov, Janice Lam, Daniel Li, Xutai Ma, Ruslan Mavlyutov, Benjamin Peloquin, Mohamed Ramadan, Abinesh Ramakrishnan, Anna Sun, Kevin Tran, Tuan Tran, Igor Tufanov, Vish Vogeti, Carleigh Wood, Yilin Yang, Bokai Yu, Pierre Andrews, Can Balioglu, Marta~R. Costa-jussà, Onur Celebi, Maha Elbayad, Cynthia Gao, Francisco Guzmán, Justine Kao, Ann Lee, Alexandre Mourachko, Juan Pino, Sravya Popuri, Christophe Ropers, Safiyyah Saleem, Holger Schwenk, Paden Tomasello, Changhan Wang, Jeff
  Wang, and Skyler Wang. 2023.
\newblock \href {https://arxiv.org/abs/2308.11596} {Seamlessm4t: Massively multilingual \& multimodal machine translation}.
\newblock \emph{Preprint}, arXiv:2308.11596.

\bibitem[{Del-Agua et~al.(2018)Del-Agua, Giménez, Sanchis, Civera, and Juan}]{8325512}
Miguel~Ángel Del-Agua, Adrià Giménez, Albert Sanchis, Jorge Civera, and Alfons Juan. 2018.
\newblock \href {https://doi.org/10.1109/TASLP.2018.2819900} {Speaker-adapted confidence measures for asr using deep bidirectional recurrent neural networks}.
\newblock \emph{IEEE/ACM Transactions on Audio, Speech, and Language Processing}, 26(7):1198--1206.

\bibitem[{Dhanjal and Singh(2024)}]{dhanjal2024comprehensive}
Amandeep~Singh Dhanjal and Williamjeet Singh. 2024.
\newblock A comprehensive survey on automatic speech recognition using neural networks.
\newblock \emph{Multimedia Tools and Applications}, 83(8):23367--23412.

\bibitem[{Duquenne et~al.(2023)Duquenne, Schwenk, and Sagot}]{duquenne2023sonar}
Paul-Ambroise Duquenne, Holger Schwenk, and Beno{\^\i}t Sagot. 2023.
\newblock Sonar: sentence-level multimodal and language-agnostic representations.
\newblock \emph{arXiv e-prints}, pages arXiv--2308.

\bibitem[{Elloumi et~al.(2018)Elloumi, Besacier, Galibert, Kahn, and Lecouteux}]{elloumi2018asrperformancepredictionunseen}
Zied Elloumi, Laurent Besacier, Olivier Galibert, Juliette Kahn, and Benjamin Lecouteux. 2018.
\newblock \href {https://arxiv.org/abs/1804.08477} {Asr performance prediction on unseen broadcast programs using convolutional neural networks}.
\newblock \emph{Preprint}, arXiv:1804.08477.

\bibitem[{Galvez et~al.(2021)Galvez, Diamos, Ciro, Cerón, Achorn, Gopi, Kanter, Lam, Mazumder, and Reddi}]{galvez2021peoplesspeechlargescalediverse}
Daniel Galvez, Greg Diamos, Juan Ciro, Juan~Felipe Cerón, Keith Achorn, Anjali Gopi, David Kanter, Maximilian Lam, Mark Mazumder, and Vijay~Janapa Reddi. 2021.
\newblock \href {https://arxiv.org/abs/2111.09344} {The people's speech: A large-scale diverse english speech recognition dataset for commercial usage}.
\newblock \emph{Preprint}, arXiv:2111.09344.

\bibitem[{Gandhi et~al.(2023)Gandhi, von Platen, and Rush}]{gandhi2023distilwhisper}
Sanchit Gandhi, Patrick von Platen, and Alexander~M. Rush. 2023.
\newblock \href {https://arxiv.org/abs/2311.00430} {Distil-whisper: Robust knowledge distillation via large-scale pseudo labelling}.
\newblock \emph{Preprint}, arXiv:2311.00430.

\bibitem[{Gulati et~al.(2020)Gulati, Qin, Chiu, Parmar, Zhang, Yu, Han, Wang, Zhang, Wu, and Pang}]{gulati2020conformerconvolutionaugmentedtransformerspeech}
Anmol Gulati, James Qin, Chung-Cheng Chiu, Niki Parmar, Yu~Zhang, Jiahui Yu, Wei Han, Shibo Wang, Zhengdong Zhang, Yonghui Wu, and Ruoming Pang. 2020.
\newblock \href {https://arxiv.org/abs/2005.08100} {Conformer: Convolution-augmented transformer for speech recognition}.
\newblock \emph{Preprint}, arXiv:2005.08100.

\bibitem[{Guo et~al.(2021)Guo, Chang, Watanabe, and Xie}]{guo2021multispeakerasrcombiningnonautoregressive}
Pengcheng Guo, Xuankai Chang, Shinji Watanabe, and Lei Xie. 2021.
\newblock \href {https://arxiv.org/abs/2106.08595} {Multi-speaker asr combining non-autoregressive conformer ctc and conditional speaker chain}.
\newblock \emph{Preprint}, arXiv:2106.08595.

\bibitem[{Harper et~al.(2024)Harper, Majumdar, Kuchaiev, Jason, Zhang, Bakhturina, Noroozi, Subramanian, Nithin, Jocelyn, Jia, Balam, Yang, Livne, Dong, Naren, and Ginsburg}]{Harper_NeMo_a_toolkit}
Eric Harper, Somshubra Majumdar, Oleksii Kuchaiev, Li~Jason, Yang Zhang, Evelina Bakhturina, Vahid Noroozi, Sandeep Subramanian, Koluguri Nithin, Huang Jocelyn, Fei Jia, Jagadeesh Balam, Xuesong Yang, Micha Livne, Yi~Dong, Sean Naren, and Boris Ginsburg. 2024.
\newblock \href {https://github.com/NVIDIA/NeMo} {{NeMo: a toolkit for Conversational AI and Large Language Models}}.

\bibitem[{Hsu et~al.(2021)Hsu, Bolte, Tsai, Lakhotia, Salakhutdinov, and Mohamed}]{hsu2021hubertselfsupervisedspeechrepresentation}
Wei-Ning Hsu, Benjamin Bolte, Yao-Hung~Hubert Tsai, Kushal Lakhotia, Ruslan Salakhutdinov, and Abdelrahman Mohamed. 2021.
\newblock \href {https://arxiv.org/abs/2106.07447} {Hubert: Self-supervised speech representation learning by masked prediction of hidden units}.
\newblock \emph{Preprint}, arXiv:2106.07447.

\bibitem[{Jalalvand et~al.(2016)Jalalvand, Negri, Turchi, C.~de Souza, Falavigna, and Qwaider}]{jalalvand-etal-2016-transcrater}
Shahab Jalalvand, Matteo Negri, Marco Turchi, Jos{\'e}~G. C.~de Souza, Daniele Falavigna, and Mohammed R.~H. Qwaider. 2016.
\newblock \href {https://doi.org/10.18653/v1/P16-4008} {{T}ransc{R}ater: a tool for automatic speech recognition quality estimation}.
\newblock In \emph{Proceedings of {ACL}-2016 System Demonstrations}, pages 43--48, Berlin, Germany. Association for Computational Linguistics.

\bibitem[{{Jan van Doorn}(2023)}]{jan_van_doorn_2023}
{Jan van Doorn}. 2023.
\newblock \href {https://doi.org/10.57967/hf/1378} {atcosim (revision b5839d9)}.

\bibitem[{Jeffries et~al.(2024)Jeffries, King, Kudlur, Nicholson, Wang, and Warden}]{jeffries2024moonshinespeechrecognitionlive}
Nat Jeffries, Evan King, Manjunath Kudlur, Guy Nicholson, James Wang, and Pete Warden. 2024.
\newblock \href {https://arxiv.org/abs/2410.15608} {Moonshine: Speech recognition for live transcription and voice commands}.
\newblock \emph{Preprint}, arXiv:2410.15608.

\bibitem[{Kalgaonkar et~al.(2015)Kalgaonkar, Liu, Gong, and Yao}]{7178922}
Kaustubh Kalgaonkar, Chaojun Liu, Yifan Gong, and Kaisheng Yao. 2015.
\newblock \href {https://doi.org/10.1109/ICASSP.2015.7178922} {Estimating confidence scores on asr results using recurrent neural networks}.
\newblock In \emph{2015 IEEE International Conference on Acoustics, Speech and Signal Processing (ICASSP)}, pages 4999--5003.

\bibitem[{Karbasi and Kolossa(2022)}]{karbasi2022asr}
Mahdie Karbasi and Dorothea Kolossa. 2022.
\newblock Asr-based speech intelligibility prediction: A review.
\newblock \emph{Hearing Research}, 426:108606.

\bibitem[{Kheddar et~al.(2024)Kheddar, Hemis, and Himeur}]{kheddar2024automatic}
Hamza Kheddar, Mustapha Hemis, and Yassine Himeur. 2024.
\newblock Automatic speech recognition using advanced deep learning approaches: A survey.
\newblock \emph{Information Fusion}, page 102422.

\bibitem[{Kuhn et~al.(2024)Kuhn, Kersken, Reuter, Egger, and Zimmermann}]{kuhn2024measuring}
Korbinian Kuhn, Verena Kersken, Benedikt Reuter, Niklas Egger, and Gottfried Zimmermann. 2024.
\newblock Measuring the accuracy of automatic speech recognition solutions.
\newblock \emph{ACM Transactions on Accessible Computing}, 16(4):1--23.

\bibitem[{Likhomanenko et~al.(2020)Likhomanenko, Xu, Pratap, Tomasello, Kahn, Avidov, Collobert, and Synnaeve}]{likhomanenko2020rethinking}
Tatiana Likhomanenko, Qiantong Xu, Vineel Pratap, Paden Tomasello, Jacob Kahn, Gilad Avidov, Ronan Collobert, and Gabriel Synnaeve. 2020.
\newblock Rethinking evaluation in asr: Are our models robust enough?
\newblock \emph{arXiv preprint arXiv:2010.11745}.

\bibitem[{Lin et~al.(2021)Lin, Zheng, Chu, Han, Hung, Ho, Chang, and Lai}]{lin2021speech}
Yu-Yi Lin, Wei-Zhong Zheng, Wei~Chung Chu, Ji-Yan Han, Ying-Hsiu Hung, Guan-Min Ho, Chia-Yuan Chang, and Ying-Hui Lai. 2021.
\newblock A speech command control-based recognition system for dysarthric patients based on deep learning technology.
\newblock \emph{Applied Sciences}, 11(6):2477.

\bibitem[{Litman et~al.(2000)Litman, Hirschberg, and Swerts}]{litman-etal-2000-predicting}
Diane~J. Litman, Julia~B. Hirschberg, and Marc Swerts. 2000.
\newblock \href {https://aclanthology.org/A00-2029/} {Predicting automatic speech recognition performance using prosodic cues}.
\newblock In \emph{1st Meeting of the North {A}merican Chapter of the Association for Computational Linguistics}.

\bibitem[{Liu et~al.(2019)Liu, Ott, Goyal, Du, Joshi, Chen, Levy, Lewis, Zettlemoyer, and Stoyanov}]{liu2019robertarobustlyoptimizedbert}
Yinhan Liu, Myle Ott, Naman Goyal, Jingfei Du, Mandar Joshi, Danqi Chen, Omer Levy, Mike Lewis, Luke Zettlemoyer, and Veselin Stoyanov. 2019.
\newblock \href {https://arxiv.org/abs/1907.11692} {Roberta: A robustly optimized bert pretraining approach}.
\newblock \emph{Preprint}, arXiv:1907.11692.

\bibitem[{Negri et~al.(2014)Negri, Turchi, C.~de Souza, and Falavigna}]{negri-etal-2014-quality}
Matteo Negri, Marco Turchi, Jos{\'e}~G. C.~de Souza, and Daniele Falavigna. 2014.
\newblock \href {https://aclanthology.org/C14-1171/} {Quality estimation for automatic speech recognition}.
\newblock In \emph{Proceedings of {COLING} 2014, the 25th International Conference on Computational Linguistics: Technical Papers}, pages 1813--1823, Dublin, Ireland. Dublin City University and Association for Computational Linguistics.

\bibitem[{Noroozi et~al.(2024)Noroozi, Majumdar, Kumar, Balam, and Ginsburg}]{noroozi2024statefulconformercachebasedinference}
Vahid Noroozi, Somshubra Majumdar, Ankur Kumar, Jagadeesh Balam, and Boris Ginsburg. 2024.
\newblock \href {https://arxiv.org/abs/2312.17279} {Stateful conformer with cache-based inference for streaming automatic speech recognition}.
\newblock \emph{Preprint}, arXiv:2312.17279.

\bibitem[{Ospanov et~al.(2024)Ospanov, Zhang, Jalali, Cao, Bogdanov, and Farnia}]{ospanov2024towards}
Azim Ospanov, Jingwei Zhang, Mohammad Jalali, Xuenan Cao, Andrej Bogdanov, and Farzan Farnia. 2024.
\newblock Towards a scalable reference-free evaluation of generative models.
\newblock \emph{arXiv preprint arXiv:2407.02961}.

\bibitem[{Panayotov et~al.(2015)Panayotov, Chen, Povey, and Khudanpur}]{panayotov2015librispeech}
Vassil Panayotov, Guoguo Chen, Daniel Povey, and Sanjeev Khudanpur. 2015.
\newblock Librispeech: An asr corpus based on public domain audio books.
\newblock \url{http://www.openslr.org/12/}.
\newblock Accessed: [Insert Date].

\bibitem[{Papadopoulos~Korfiatis et~al.(2022)Papadopoulos~Korfiatis, Moramarco, Sarac, and Savkov}]{papadopoulos-korfiatis-etal-2022-primock57}
Alex Papadopoulos~Korfiatis, Francesco Moramarco, Radmila Sarac, and Aleksandar Savkov. 2022.
\newblock \href {https://doi.org/10.18653/v1/2022.acl-short.65} {{P}ri{M}ock57: A dataset of primary care mock consultations}.
\newblock In \emph{Proceedings of the 60th Annual Meeting of the Association for Computational Linguistics (Volume 2: Short Papers)}, pages 588--598, Dublin, Ireland. Association for Computational Linguistics.

\bibitem[{Park et~al.(2024)Park, Kang, and Hain}]{park2024character}
Chanho Park, Hyunsik Kang, and Thomas Hain. 2024.
\newblock Character error rate estimation for automatic speech recognition of short utterances.
\newblock In \emph{2024 32nd European Signal Processing Conference (EUSIPCO)}, pages 131--135. IEEE.

\bibitem[{Pratap et~al.(2023)Pratap, Tjandra, Shi, Tomasello, Babu, Kundu, Elkahky, Ni, Vyas, Fazel-Zarandi, Baevski, Adi, Zhang, Hsu, Conneau, and Auli}]{pratap2023scalingspeechtechnology1000}
Vineel Pratap, Andros Tjandra, Bowen Shi, Paden Tomasello, Arun Babu, Sayani Kundu, Ali Elkahky, Zhaoheng Ni, Apoorv Vyas, Maryam Fazel-Zarandi, Alexei Baevski, Yossi Adi, Xiaohui Zhang, Wei-Ning Hsu, Alexis Conneau, and Michael Auli. 2023.
\newblock \href {https://arxiv.org/abs/2305.13516} {Scaling speech technology to 1,000+ languages}.
\newblock \emph{Preprint}, arXiv:2305.13516.

\bibitem[{Qiu et~al.(2021)Qiu, Li, He, Zhang, Li, Cao, Prabhavalkar, Bhatia, Li, Hu, Sainath, and McGraw}]{9413966}
David Qiu, Qiujia Li, Yanzhang He, Yu~Zhang, Bo~Li, Liangliang Cao, Rohit Prabhavalkar, Deepti Bhatia, Wei Li, Ke~Hu, Tara~N. Sainath, and Ian McGraw. 2021.
\newblock \href {https://doi.org/10.1109/ICASSP39728.2021.9413966} {Learning word-level confidence for subword end-to-end asr}.
\newblock In \emph{ICASSP 2021 - 2021 IEEE International Conference on Acoustics, Speech and Signal Processing (ICASSP)}, pages 6393--6397.

\bibitem[{Radford et~al.(2022)Radford, Kim, Xu, Brockman, McLeavey, and Sutskever}]{radford2022robustspeechrecognitionlargescale}
Alec Radford, Jong~Wook Kim, Tao Xu, Greg Brockman, Christine McLeavey, and Ilya Sutskever. 2022.
\newblock \href {https://arxiv.org/abs/2212.04356} {Robust speech recognition via large-scale weak supervision}.
\newblock \emph{Preprint}, arXiv:2212.04356.

\bibitem[{Radford et~al.(2023)Radford, Kim, Xu, Brockman, McLeavey, and Sutskever}]{radford2023robust}
Alec Radford, Jong~Wook Kim, Tao Xu, Greg Brockman, Christine McLeavey, and Ilya Sutskever. 2023.
\newblock Robust speech recognition via large-scale weak supervision.
\newblock In \emph{International conference on machine learning}, pages 28492--28518. PMLR.

\bibitem[{Raj et~al.(2011)Raj, Singh, and Baker}]{5947648}
Bhiksha Raj, Rita Singh, and James Baker. 2011.
\newblock \href {https://doi.org/10.1109/ICASSP.2011.5947648} {A paired test for recognizer selection with untranscribed data}.
\newblock In \emph{2011 IEEE International Conference on Acoustics, Speech and Signal Processing (ICASSP)}, pages 5676--5679.

\bibitem[{Rajaa and Tushar()}]{Rajaa_SpeechLLM_Multi-Modal_LLM}
Shangeth Rajaa and Abhinav Tushar.
\newblock \href {https://github.com/skit-ai/SpeechLLM} {{SpeechLLM: Multi-Modal LLM for Speech Understanding}}.

\bibitem[{Rekesh et~al.(2023)Rekesh, Koluguri, Kriman, Majumdar, Noroozi, Huang, Hrinchuk, Puvvada, Kumar, Balam, and Ginsburg}]{rekesh2023fastconformerlinearlyscalable}
Dima Rekesh, Nithin~Rao Koluguri, Samuel Kriman, Somshubra Majumdar, Vahid Noroozi, He~Huang, Oleksii Hrinchuk, Krishna Puvvada, Ankur Kumar, Jagadeesh Balam, and Boris Ginsburg. 2023.
\newblock \href {https://arxiv.org/abs/2305.05084} {Fast conformer with linearly scalable attention for efficient speech recognition}.
\newblock \emph{Preprint}, arXiv:2305.05084.

\bibitem[{Rio et~al.(2022)Rio, Ha, McNamara, Miller, and Chandra}]{rio2022earnings}
MD~Rio, Peter Ha, Quinten McNamara, Corey Miller, and Shipra Chandra. 2022.
\newblock Earnings-22: A practical benchmark for accents in the wild.

\bibitem[{Rousseau et~al.(2014)Rousseau, Del{\'e}glise, and Est{\`e}ve}]{rousseau2014tedlium}
Anthony Rousseau, Paul Del{\'e}glise, and Yann Est{\`e}ve. 2014.
\newblock Ted-lium 3: Twice as much data and corpus repartition for experiments on speaker adaptation.
\newblock \url{https://lium.univ-lemans.fr/ted-lium3/}.
\newblock Accessed: [Insert Date].

\bibitem[{Schneider et~al.(2019)Schneider, Baevski, Collobert, and Auli}]{schneider2019wav2vecunsupervisedpretrainingspeech}
Steffen Schneider, Alexei Baevski, Ronan Collobert, and Michael Auli. 2019.
\newblock \href {https://arxiv.org/abs/1904.05862} {wav2vec: Unsupervised pre-training for speech recognition}.
\newblock \emph{Preprint}, arXiv:1904.05862.

\bibitem[{{Seamless Communication} et~al.(2023){Seamless Communication}, Barrault, Chung, Meglioli, Dale, Dong, Duppenthaler, Duquenne, Ellis, Elsahar, Haaheim, Hoffman, Hwang, Inaguma, Klaiber, Kulikov, Li, Licht, Maillard, Mavlyutov, Rakotoarison, Sadagopan, Ramakrishnan, Tran, Wenzek, Yang, Ye, Evtimov, Fernandez, Gao, Hansanti, Kalbassi, Kallet, Kozhevnikov, Mejia, Roman, Touret, Wong, Wood, Yu, Andrews, Balioglu, Chen, Costa-juss{\`a}, Elbayad, Gong, Guzm{\'a}n, Heffernan, Jain, Kao, Lee, Ma, Mourachko, Peloquin, Pino, Popuri, Ropers, Saleem, Schwenk, Sun, Tomasello, Wang, Wang, Wang, and Williamson}]{seamless2023}
{Seamless Communication}, Lo{\"i}c Barrault, Yu-An Chung, Mariano~Coria Meglioli, David Dale, Ning Dong, Mark Duppenthaler, Paul-Ambroise Duquenne, Brian Ellis, Hady Elsahar, Justin Haaheim, John Hoffman, Min-Jae Hwang, Hirofumi Inaguma, Christopher Klaiber, Ilia Kulikov, Pengwei Li, Daniel Licht, Jean Maillard, Ruslan Mavlyutov, Alice Rakotoarison, Kaushik~Ram Sadagopan, Abinesh Ramakrishnan, Tuan Tran, Guillaume Wenzek, Yilin Yang, Ethan Ye, Ivan Evtimov, Pierre Fernandez, Cynthia Gao, Prangthip Hansanti, Elahe Kalbassi, Amanda Kallet, Artyom Kozhevnikov, Gabriel Mejia, Robin~San Roman, Christophe Touret, Corinne Wong, Carleigh Wood, Bokai Yu, Pierre Andrews, Can Balioglu, Peng-Jen Chen, Marta~R. Costa-juss{\`a}, Maha Elbayad, Hongyu Gong, Francisco Guzm{\'a}n, Kevin Heffernan, Somya Jain, Justine Kao, Ann Lee, Xutai Ma, Alex Mourachko, Benjamin Peloquin, Juan Pino, Sravya Popuri, Christophe Ropers, Safiyyah Saleem, Holger Schwenk, Anna Sun, Paden Tomasello, Changhan Wang, Jeff Wang, Skyler Wang, and Mary
  Williamson. 2023.
\newblock Seamless: Multilingual expressive and streaming speech translation.

\bibitem[{Sheshadri et~al.(2021{\natexlab{a}})Sheshadri, Rao~Vijjini, and Kharbanda}]{sheshadri-etal-2021-wer}
Akshay~Krishna Sheshadri, Anvesh Rao~Vijjini, and Sukhdeep Kharbanda. 2021{\natexlab{a}}.
\newblock \href {https://doi.org/10.18653/v1/2021.eacl-main.320} {{WER}-{BERT}: Automatic {WER} estimation with {BERT} in a balanced ordinal classification paradigm}.
\newblock In \emph{Proceedings of the 16th Conference of the European Chapter of the Association for Computational Linguistics: Main Volume}, pages 3661--3672, Online. Association for Computational Linguistics.

\bibitem[{Sheshadri et~al.(2021{\natexlab{b}})Sheshadri, Vijjini, and Kharbanda}]{sheshadri2021werbertautomaticwerestimation}
Akshay~Krishna Sheshadri, Anvesh~Rao Vijjini, and Sukhdeep Kharbanda. 2021{\natexlab{b}}.
\newblock \href {https://arxiv.org/abs/2101.05478} {Wer-bert: Automatic wer estimation with bert in a balanced ordinal classification paradigm}.
\newblock \emph{Preprint}, arXiv:2101.05478.

\bibitem[{Shon et~al.(2022)Shon, Pasad, Wu, Brusco, Artzi, Livescu, and Han}]{shon2022slue}
Suwon Shon, Ankita Pasad, Felix Wu, Pablo Brusco, Yoav Artzi, Karen Livescu, and Kyu~J Han. 2022.
\newblock Slue: New benchmark tasks for spoken language understanding evaluation on natural speech.
\newblock In \emph{ICASSP 2022-2022 IEEE International Conference on Acoustics, Speech and Signal Processing (ICASSP)}, pages 7927--7931. IEEE.

\bibitem[{Swarup et~al.(2019)Swarup, Maas, Garimella, Mallidi, and Hoffmeister}]{Swarup2019}
Prakhar Swarup, Roland Maas, Sri Garimella, Sri~Harish Mallidi, and Björn Hoffmeister. 2019.
\newblock \href {https://www.amazon.science/publications/improving-asr-confidence-scores-for-alexa-using-acoustic-and-hypothesis-embeddings} {Improving asr confidence scores for alexa using acoustic and hypothesis embeddings}.

\bibitem[{Tang et~al.(2023)Tang, Sun, Inaguma, Chen, Dong, Ma, Tomasello, and Pino}]{tang2023hybridtransducerattentionbased}
Yun Tang, Anna~Y. Sun, Hirofumi Inaguma, Xinyue Chen, Ning Dong, Xutai Ma, Paden~D. Tomasello, and Juan Pino. 2023.
\newblock \href {https://arxiv.org/abs/2305.03101} {Hybrid transducer and attention based encoder-decoder modeling for speech-to-text tasks}.
\newblock \emph{Preprint}, arXiv:2305.03101.

\bibitem[{Technologies(2021)}]{kensho2021spgispeech}
Kensho Technologies. 2021.
\newblock Spgispeech: A large-scale, high-quality dataset for speech recognition in financial earnings calls.
\newblock \url{https://datasets.kensho.com/datasets/spgispeech}.
\newblock Accessed: [Insert Date].

\bibitem[{Tuttösí et~al.(2025)Tuttösí, Dhillon, Sang, Eastwood, Bhatia, Dinh, Kapoor, Jin, and Lim}]{tuttösí2025berstingscreamsbenchmarkdistanced}
Paige Tuttösí, Mantaj Dhillon, Luna Sang, Shane Eastwood, Poorvi Bhatia, Quang~Minh Dinh, Avni Kapoor, Yewon Jin, and Angelica Lim. 2025.
\newblock \href {https://arxiv.org/abs/2505.00059} {Bersting at the screams: A benchmark for distanced, emotional and shouted speech recognition}.
\newblock \emph{Preprint}, arXiv:2505.00059.

\bibitem[{Variani et~al.(2020)Variani, Rybach, Allauzen, and Riley}]{variani2020hybridautoregressivetransducerhat}
Ehsan Variani, David Rybach, Cyril Allauzen, and Michael Riley. 2020.
\newblock \href {https://arxiv.org/abs/2003.07705} {Hybrid autoregressive transducer (hat)}.
\newblock \emph{Preprint}, arXiv:2003.07705.

\bibitem[{Vaswani et~al.(2023)Vaswani, Shazeer, Parmar, Uszkoreit, Jones, Gomez, Kaiser, and Polosukhin}]{vaswani2023attentionneed}
Ashish Vaswani, Noam Shazeer, Niki Parmar, Jakob Uszkoreit, Llion Jones, Aidan~N. Gomez, Lukasz Kaiser, and Illia Polosukhin. 2023.
\newblock \href {https://arxiv.org/abs/1706.03762} {Attention is all you need}.
\newblock \emph{Preprint}, arXiv:1706.03762.

\bibitem[{Waheed et~al.(2024)Waheed, Kadaoui, and Abdul-Mageed}]{waheed-etal-2024-distill}
Abdul Waheed, Karima Kadaoui, and Muhammad Abdul-Mageed. 2024.
\newblock \href {https://doi.org/10.18653/v1/2024.acl-long.680} {To distill or not to distill? on the robustness of robust knowledge distillation}.
\newblock In \emph{Proceedings of the 62nd Annual Meeting of the Association for Computational Linguistics (Volume 1: Long Papers)}, pages 12603--12621, Bangkok, Thailand. Association for Computational Linguistics.

\bibitem[{Waheed et~al.(2025)Waheed, Kadaoui, Raj, and Abdul-Mageed}]{waheed2025udistilwhisperlabelfreedatafiltering}
Abdul Waheed, Karima Kadaoui, Bhiksha Raj, and Muhammad Abdul-Mageed. 2025.
\newblock \href {https://arxiv.org/abs/2407.01257} {udistil-whisper: Label-free data filtering for knowledge distillation in low-data regimes}.
\newblock \emph{Preprint}, arXiv:2407.01257.

\bibitem[{Wang et~al.(2021)Wang, Riviere, Lee, Wu, Talnikar, Haziza, Williamson, Pino, and Dupoux}]{wang-etal-2021-voxpopuli}
Changhan Wang, Morgane Riviere, Ann Lee, Anne Wu, Chaitanya Talnikar, Daniel Haziza, Mary Williamson, Juan Pino, and Emmanuel Dupoux. 2021.
\newblock \href {https://aclanthology.org/2021.acl-long.80} {{V}ox{P}opuli: A large-scale multilingual speech corpus for representation learning, semi-supervised learning and interpretation}.
\newblock In \emph{Proceedings of the 59th Annual Meeting of the Association for Computational Linguistics and the 11th International Joint Conference on Natural Language Processing (Volume 1: Long Papers)}, pages 993--1003, Online. Association for Computational Linguistics.

\bibitem[{Wang et~al.(2024)Wang, Edraki, Chan, L{\'o}pez-Espejo, and Jensen}]{wang2024no}
Haolan Wang, Amin Edraki, Wai-Yip Chan, Iv{\'a}n L{\'o}pez-Espejo, and Jesper Jensen. 2024.
\newblock No-reference speech intelligibility prediction leveraging a noisy-speech asr pre-trained model.
\newblock In \emph{Proc. Interspeech 2024}, pages 3849--3853.

\bibitem[{Watanabe et~al.(2020)Watanabe, Mandel, Barker, Vincent, Arora, Chang, Khudanpur, Manohar, Povey, Raj et~al.}]{watanabe2020chime}
Shinji Watanabe, Michael Mandel, Jon Barker, Emmanuel Vincent, Ashish Arora, Xuankai Chang, Sanjeev Khudanpur, Vimal Manohar, Daniel Povey, Desh Raj, et~al. 2020.
\newblock Chime-6 challenge: Tackling multispeaker speech recognition for unsegmented recordings.
\newblock In \emph{CHiME 2020-6th International Workshop on Speech Processing in Everyday Environments}.

\bibitem[{Wolf et~al.(2020)Wolf, Debut, Sanh, Chaumond, Delangue, Moi, Cistac, Rault, Louf, Funtowicz, Davison, Shleifer, von Platen, Ma, Jernite, Plu, Xu, Scao, Gugger, Drame, Lhoest, and Rush}]{wolf2020huggingfacestransformersstateoftheartnatural}
Thomas Wolf, Lysandre Debut, Victor Sanh, Julien Chaumond, Clement Delangue, Anthony Moi, Pierric Cistac, Tim Rault, Rémi Louf, Morgan Funtowicz, Joe Davison, Sam Shleifer, Patrick von Platen, Clara Ma, Yacine Jernite, Julien Plu, Canwen Xu, Teven~Le Scao, Sylvain Gugger, Mariama Drame, Quentin Lhoest, and Alexander~M. Rush. 2020.
\newblock \href {https://arxiv.org/abs/1910.03771} {Huggingface's transformers: State-of-the-art natural language processing}.
\newblock \emph{Preprint}, arXiv:1910.03771.

\bibitem[{Yoon et~al.(2010)Yoon, Chen, and Zechner}]{yoon10b_interspeech}
Su-Youn Yoon, Lei Chen, and Klaus Zechner. 2010.
\newblock \href {https://doi.org/10.21437/Interspeech.2010-282} {Predicting word accuracy for the automatic speech recognition of non-native speech}.
\newblock In \emph{Interspeech 2010}, pages 773--776.

\bibitem[{Yuksel et~al.(2023{\natexlab{a}})Yuksel, Ferreira, Gunduz, Al-Badrashiny, and Javadi}]{yuksel2023referencelessqualitymetricautomatic}
Kamer~Ali Yuksel, Thiago Ferreira, Ahmet Gunduz, Mohamed Al-Badrashiny, and Golara Javadi. 2023{\natexlab{a}}.
\newblock \href {https://arxiv.org/abs/2306.13114} {A reference-less quality metric for automatic speech recognition via contrastive-learning of a multi-language model with self-supervision}.
\newblock \emph{Preprint}, arXiv:2306.13114.

\bibitem[{Yuksel et~al.(2023{\natexlab{b}})Yuksel, Ferreira, Javadi, El-Badrashiny, and Gunduz}]{yuksel2023noreferreferencelessqualitymetric}
Kamer~Ali Yuksel, Thiago Ferreira, Golara Javadi, Mohamed El-Badrashiny, and Ahmet Gunduz. 2023{\natexlab{b}}.
\newblock \href {https://arxiv.org/abs/2306.12577} {Norefer: a referenceless quality metric for automatic speech recognition via semi-supervised language model fine-tuning with contrastive learning}.
\newblock \emph{Preprint}, arXiv:2306.12577.

\bibitem[{Zimerman and Wolf(2023)}]{zimerman2023long}
Itamar Zimerman and Lior Wolf. 2023.
\newblock On the long range abilities of transformers.
\newblock \emph{arXiv preprint arXiv:2311.16620}.

\end{thebibliography}
